\documentclass{article}
\usepackage{amsmath,amssymb}
\usepackage{graphicx}
\usepackage{xcolor}
\usepackage{placeins}
\usepackage{colortbl}

\usepackage{microtype}
\usepackage{graphicx}

\usepackage{booktabs} % for professional tables
\usepackage{hyperref}
\usepackage{wrapfig}
\usepackage{soul}
\usepackage[numbers]{natbib}
\usepackage{authblk}
\usepackage[T1]{fontenc}    % use 8-bit T1 fonts
\usepackage{nicefrac}       % compact symbols for 1/2, etc.
\usepackage{amsmath,amsthm}
\usepackage{amssymb}
\usepackage{mathtools}
\usepackage{algorithm}
\usepackage{color}
\usepackage{subcaption}
\usepackage{fullpage}
\usepackage{algpseudocode}
\usepackage{bbm}
\usepackage{caption}

\usepackage[capitalize,noabbrev]{cleveref}

\theoremstyle{plain}

\theoremstyle{definition}

\theoremstyle{remark}

\usepackage[utf8]{inputenc} % allow utf-8 input
\usepackage{url}            % simple URL typesetting
\usepackage{amsfonts}       % blackboard math symbols
\usepackage{xcolor}         % colors

\usepackage{color-edits}
\addauthor{st}{blue}
\addauthor{lpf}{green}

\usepackage{subfiles}

\title{Stuck in the Matrix: Probing Spatial Reasoning \\in Large Language Models}
\author{Maggie Bai, Ava Kim Cohen, Eleanor Koss, Charlie Lichtenbaum\footnote{Correspondence to: Charlie Lichtenbaum (clichtenbaum@gmail.com)}  \footnote{Alphabetical ordering, all authors contributed equally}}

\begin{document}
    \maketitle

\begin{abstract}
This paper explores the spatial reasoning capability of large language models (LLMs) over textual input through a suite of five tasks aimed at probing their spatial understanding and computational abilities. The models were tested on both fundamental spatial reasoning and multi-step problem-solving within structured grid-based environments using tasks such as quadrant identification, geometric transformations, distance evaluation, word searches, and tile sliding. Each task was scaled in complexity through increasing grid dimensions, requiring models to extend beyond simple pattern recognition into abstract spatial reasoning. Our results reveal that while LLMs demonstrate moderate success in all tasks with small complexity and size, performance drops off rapidly as scale increases, with an average loss in accuracy of 42.7\%, and reaching as high as 84\%. Every test that began with over 50\% accuracy showed a loss of at least 48\%, illustrating the consistent nature of the deterioration. Furthermore, their struggles with scaling complexity hint at a lack of robust spatial representations in their underlying architectures. This paper underscores the gap between linguistic and spatial reasoning in LLMs, offering insights into their current limitations, and laying the groundwork for future integrative benchmarks at the intersection of language and geometry.

\end{abstract}

\section{Introduction}
Large language models (LLMs) have achieved remarkable success across a wide array of language-related tasks, including text generation, summarization, and question answering \cite{brown2020} \cite{achaim2023}, and many benchmarks, such as TruthfulQA \cite{lin2021}, have been developed to evaluate this performance. Their ability to handle non-linguistic reasoning tasks, particularly those involving geometry and spatial relationships, has received significant attention from research into Visual Language Models (VLMs), both on training methods \cite{SpatialVLM} and benchmarks \cite{MindTheGap}. Yet less work has been done on the intersection of the two, namely an LLM's spatial reasoning ability over text-based inputs. 

Moreover, we observed in attempting to have LLMs play simple games like 2048 or Connect 4 through prompt-based board inputs that while models understood the strategy and gameplay, they often still made incorrect or invalid moves due to a failure to comprehend the board state. This manifested both in the initial reading of the board and in the application of principles such as gravity in Connect 4, emphasizing the limitations of text-based spatial reasoning in LLMs. As such, while the issues could have been solved by switching to a VLM, we decided to investigate the extent of LLMs' spatial reasoning abilities in text-based domains. These capabilities may improve LLM usage and reliability in layout-heavy fields such as finance, where interpreting tabular data correctly is necessary. 

In this study, we designed a series of simple, isolated tests to measure these capabilities, with each task progressively increasing in complexity. These tasks include:

\begin{enumerate}
    \item Quadrant: determining which quadrant of a Cartesian grid contains a specific symbol.

    \item Transformation: reflecting a symbol across axes and producing the correct spatial configuration.

    \item Distance: identifying the closest and farthest symbols relative to a target point on a grid.

    \item Word Search: locating a defined word within a grid

    \item Slide: sliding an X in a cardinal direction on a grid until it hits the edge or a wall
\end{enumerate}

We evaluated four LLMs: GPT-4o, GPT-4.1, Claude 3.7 Sonnet with no thinking tokens (referred to as No Thinking), and Claude 3.7 with 16,000 thinking tokens (referred to as Medium Thinking). Each task was applied to progressively larger grid sizes, enabling detailed analysis of how well models generalize simple spatial reasoning to larger structures and more intricate configurations.

\section{Related Works}
This paper was inspired by work done with LLMs on text-based games, where experiments showed that models were unable to form and maintain the game worlds \cite{Tsai}. In addition, in Topaskal et al. \cite{topsakal2024}, models were presented with boards for Tic-Tac-Toe, Gomoku, and Connect 4 in image and text-based formats. Their results showed a higher rate of invalid moves on more complex grid states in both the image and ``illustration," or ASCII grid, formats, implying that the LLMs failed to comprehend the board properly. As such, this study sought to identify the complexity at which models begin to deteriorate as well as the types of problems the models encounter. 

LLMs are also frequently benchmarked on their geometric abilities. Datasets of math problems, such as MATH \cite{Hendrycks} or GPSM4K \cite{Anand}, have been used to test the multi-step reasoning and geometry capabilities of various LLMs. These studies have all found the space for significant improvement in LLM spatial processing and reasoning. 

However, these studies differ from ours in two significant ways. Firstly, this study was not designed to push the limits of the mathematical capabilities of the LLMs, unlike the datasets of Olympiad or graduate-level novel geometry problems. Rather, it tests basic reasoning and understanding of relative placements, with only simple arithmetic involved. And secondly, the models in our study receive their inputs in the form of ASCII grids, as opposed to images or textual descriptions of layouts. As such, this study serves better as an inquiry into tokenization and interpretation by the models.

Another benchmark in LLM spatial reasoning is the competition ARC-AGI \cite{ARC-AGI}, which has spurred numerous developments in LLM spatial reasoning capabilities over the six years since its release. The recently released ARC-AGI-2 \cite{ARC-2} will likely continue to push researchers towards further developments. This paper, however, was not largely focused on developing novel strategies for enhancing reasoning capabilities, but rather benchmarking pre-existing ones.

Finally, a similar problem to spatial reasoning is that of tabular reasoning, or the interpretation of mixed text and layout documents. This kind of reasoning is common to fields such as finance or business, where contracts and invoices often contain tabular data. Researchers have developed specific tabular models, such as DocLLM \cite{Doc}, to fit these tasks, and have shown significant improvement compared to standard LLMs. This is a similar problem to that of spatial reasoning, as understanding the structure and relative positioning of elements in a layout is imperative to accuracy, and a parsing mistake can cost a company thousands of dollars. As such, improving the spatial reasoning capabilities and reliability of LLMs could increase their use cases significantly.

\section{Methodology}
To evaluate the spatial reasoning capabilities of the models, we designed five distinct tests that each targeted a specific aspect of spatial understanding. All tasks used square grids composed of simple symbols (for example, `$\cdot$' or letters) and were progressively scaled to increase complexity. All grids were presented with spaces as delimeters between indices. An example is shown below; all grids tokenized in the same manner, just with alternate characters instead of ``."s.

\begin{figure}[h]
    \centering
    \includegraphics[height=2cm, keepaspectratio]{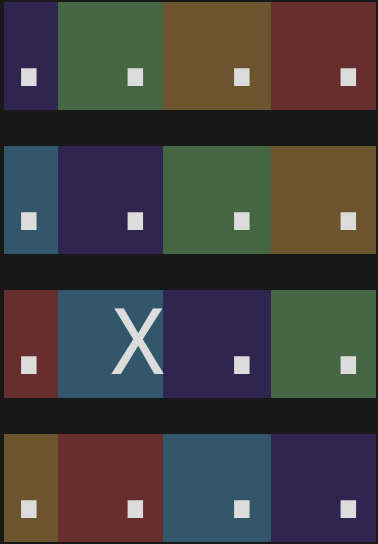}
    \caption{Example Tokenization}
\end{figure}

\subsection{Quadrant}

In this task, a single `X' was placed randomly on a grid, and the models were asked to determine the quadrant of the grid in which the `X' was located. The quadrants were defined in alignment with the Cartesian coordinate system and the models replied in a ``Lower/Upper Left/Right" format. An example grid is shown below:

\begin{figure}[h]
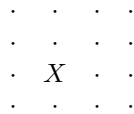

        \centering
        \[
        \begin{matrix}
        \cdot & \cdot & \cdot & \cdot\\
        \cdot & \cdot & \cdot & \cdot\\
        \cdot & X & \cdot & \cdot\\
        \cdot & \cdot & \cdot & \cdot
        \end{matrix}
    \]
        \caption{In this example, the `X' is in quadrant III (lower left quadrant).}
\end{figure}

\subsection{Transformation}

This task tested the ability of the models to perform geometric transformations. A single ``X" was randomly placed on a square grid, and the models were asked to reflect the ``X" over a centerline (not given), either horizontally or vertically, and return a coordinate pair in $(x,y)$ format. Example reflections are shown below:

\begin{figure}[htbp]
        \centering
        \begin{subfigure}{\textwidth}
        \centering
        $\begin{matrix}
            \cdot & \cdot & \cdot & \cdot\\
            X & \cdot & \cdot & \cdot\\
            \cdot & \cdot & \cdot & \cdot\\
            \cdot & \cdot & \cdot & \cdot
        \end{matrix}
        \longrightarrow
        \begin{matrix}
            \cdot & \cdot & \cdot & \cdot\\
            \cdot & \cdot & \cdot & \cdot\\
            X & \cdot & \cdot & \cdot\\
            \cdot & \cdot & \cdot & \cdot
        \end{matrix}$
        \caption{Reflection vertically.}
        \label{fig:x_reflection}
        \end{subfigure}

    % Reflection across the y-axis
\begin{subfigure}{\textwidth}
        \centering
        $\begin{matrix}
            \cdot & \cdot & \cdot & \cdot\\
            \cdot & \cdot & X & \cdot\\
            \cdot & \cdot & \cdot & \cdot\\
            \cdot & \cdot & \cdot & \cdot
        \end{matrix}
        \longrightarrow
        \begin{matrix}
            \cdot & \cdot & \cdot & \cdot\\
            \cdot & X & \cdot & \cdot\\
            \cdot & \cdot & \cdot & \cdot\\
            \cdot & \cdot & \cdot & \cdot
        \end{matrix}$
        \caption{Reflection horizontally.}
        \label{fig:y_reflection}
\end{subfigure}
\end{figure}

\subsection{Distance}

In this task, letters A-G and an X were placed on a square grid. The models were asked to identify which letter was closest to and farthest from the X. An example grid is shown below:

\begin{center}
    $\begin{matrix}
        \cdot & \cdot & B & \cdot & \cdot\\
        \cdot & F & \cdot & \cdot & A\\
        \cdot & C & \cdot & \cdot & \cdot\\
        D & \cdot & \cdot & G & \cdot\\
        E & \cdot & \cdot & \cdot & X
    \end{matrix}$
\end{center}

In this example, the closest letter is G, and the farthest is B.

\subsection{Word Search}
In this task, LLMs were provided a programmatically generated grid containing a single word along with the word contained. Models were tasked with producing a list of coordinates identifying the word's location. Credit was given for partial spellings and incorrect orderings of coordinates. An example is shown below:

\begin{center}
    $\begin{matrix}
        D & A & E & Z & A\\
        E & D & B & V & D\\
        R & A & T & M & M\\
        L & T & C & Z & Y\\
        J & J & N & N & L
    \end{matrix}$
\end{center}

In this example, the word ``RAT" is embedded horizontally in the third row.

\subsection{Slide}
In this task, models were given a grid with an X and numerous \#s. The model was instructed to `slide' the X until it hit either a wall (\#) or the edge of the grid and output the final coordinates. For multi-slide problems, credit was given for correct secondary slides, even if the starting position was affected from an earlier slide. An example grid is show below:
\begin{figure}[htbp]
\begin{center}
    $\begin{matrix}
        \cdot & \# & \# & \cdot & \cdot\\
        \cdot & \# & \cdot & X & \cdot\\
        \cdot & \cdot & \cdot & \cdot & \cdot\\
        \cdot & \# & \# & \# & \cdot\\
        \# & \cdot & \cdot & \cdot & \#
    \end{matrix}
    \longrightarrow
    \begin{matrix}
        \cdot & \# & \# & \cdot & \cdot\\
        \cdot & \# & \cdot & \cdot & \cdot\\
        \cdot & \cdot & \cdot & X & \cdot\\
        \cdot & \# & \# & \# & \cdot\\
        \# & \cdot & \cdot & \cdot & \#
    \end{matrix}
    \longrightarrow
    \begin{matrix}
        \cdot & \# & \# & \cdot & \cdot\\
        \cdot & \# & \cdot & \cdot & \cdot\\
        X & \cdot & \cdot & \cdot & \cdot\\
        \cdot & \# & \# & \# & \cdot\\
        \# & \cdot & \cdot & \cdot & \#
    \end{matrix}$
\end{center}
\caption{Slide down, then left}
\end{figure}

In this example, the X would first move to row 2, column 3 and then end at row 2, column 0.

\section{Results}
Each of the designed tasks was tested on four models: GPT-4o \cite{4o}, GPT-4.1 \cite{4.1}, Claude 3.7 No Thinking, and Claude 3.7 Medium Thinking \cite{Anthropic}. Each set of parameters (i.e. grid size, model) was run 10 times, and the results were averaged. Test prompt inputs ranged from $\sim$400 to $\sim$90,000 tokens, well below the context limit for any of the models.

Additionally, we explored some common sources of error and attempted to mitigate them. These attempts are also seen below.

\subsection{Quadrant}

\begin{table}[h!]
    \centering
    \begin{tabular}{|c|c|c|}
    \hline
        Models & Smallest 5 Grids & Largest 5 Grids \\
        \hline
        GPT-4o & 80\% & 22\% \\
        \hline
        GPT-4.1 & 86\% & 35\% \\
        \hline
        Claude No Thinking & 98\% & 61\% \\
        \hline
        Claude Medium Thinking & 98\% & 58\% \\
        \hline
    \end{tabular}
    \caption{Model Accuracy on Small and Large Grids (Quadrant)}
    \label{tab:quadrant}
\end{table}

\FloatBarrier

\begin{figure}[h!]
    \centering
    \includegraphics[width=0.55\linewidth]{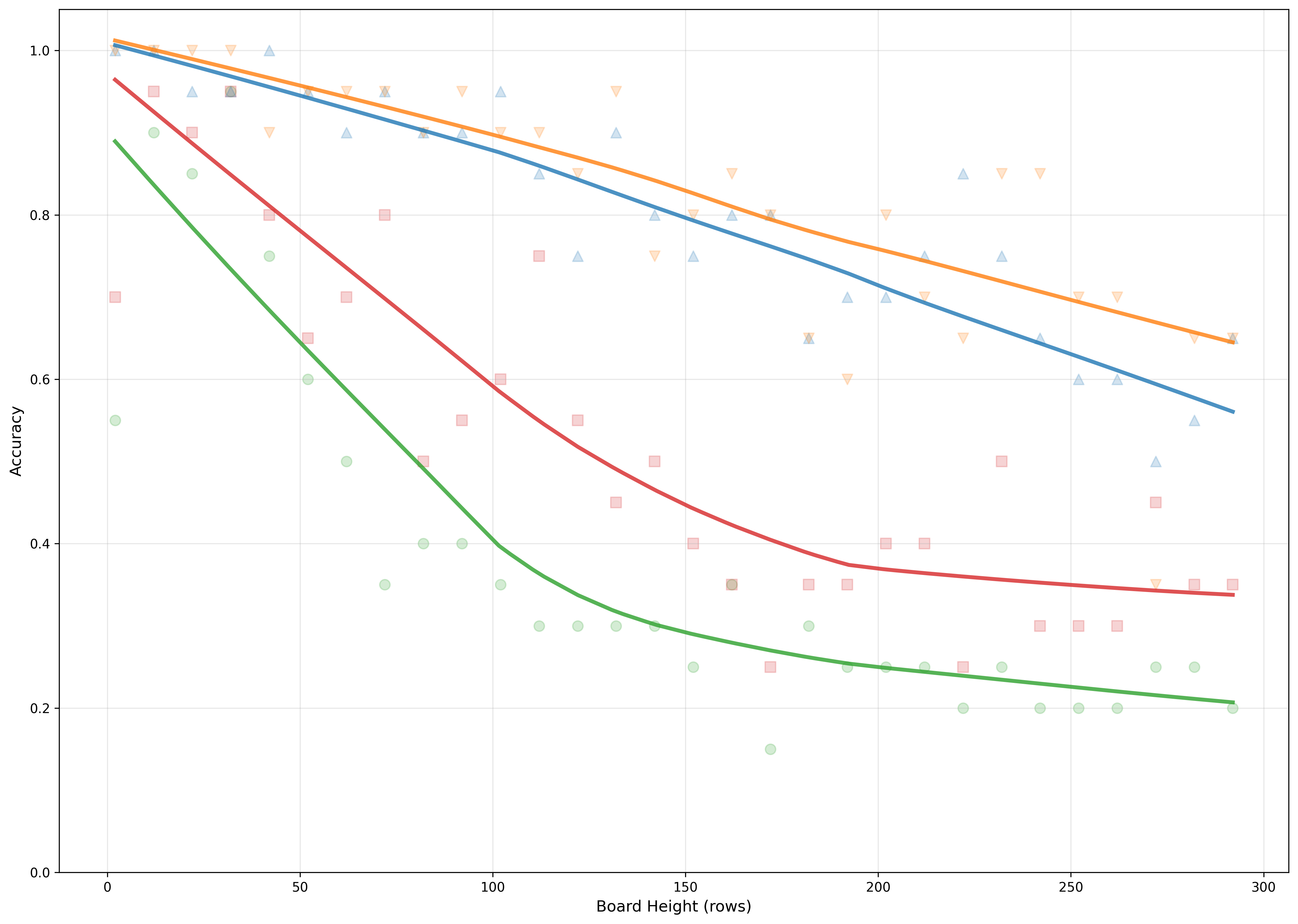}
    \caption{Accuracy vs Grid Size (Quadrant) }
    \label{fig:quad_accuracy}
\end{figure}

Grid sizes ranged from 2×2 to 300$\times$300 in increments of 10$\times$10. Small grids saw high accuracy across all models, but performance deteriorated significantly on larger grids. Specifically, the OpenAI variants performed poorly on larger grids, with accuracy dropping to 20-30\%, and Anthropic models dropping to around 60\%, as seen in Figure~\ref{fig:quad_accuracy}.

\begin{figure}[h!]
    \centering
    \includegraphics[width=0.55\linewidth]{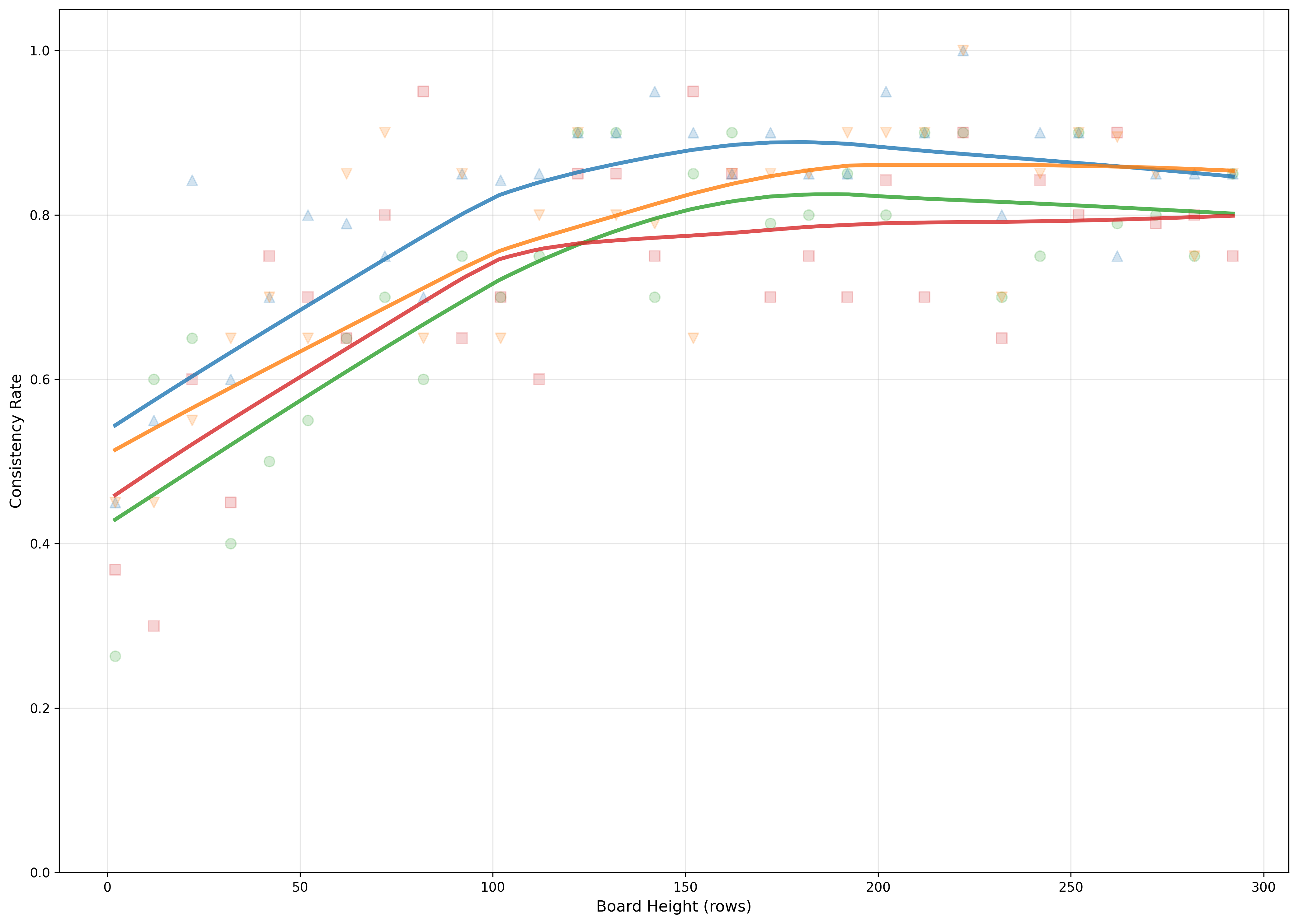}
    \caption{Consistency vs Grid Size (Quadrant)}
    \label{fig:quad_consistency}
\end{figure}

Error analysis revealed that while initial misreading of the grid was common, models were largely consistent in quadrant selection based on their internal understanding of the board state. That is to say, once the model had ``found" the X in the grid, it predicted the correct quadrant for that coordinate position; errors appeared largely when the model identified the incorrect position as containing the X. This trend is shown in Figure~\ref{fig:quad_consistency}.

The models were internally inconsistent only when the X was near a dividing line between two quadrants. Even when given the board size explicitly, models often had a difficult time establishing the location of the center lines. The increase in consistency at larger board sizes can likely be explained by the lower chance of an X placement near one of the dividing lines. 

\begin{figure}[h!]
    \centering
    \includegraphics[width=0.5\linewidth]{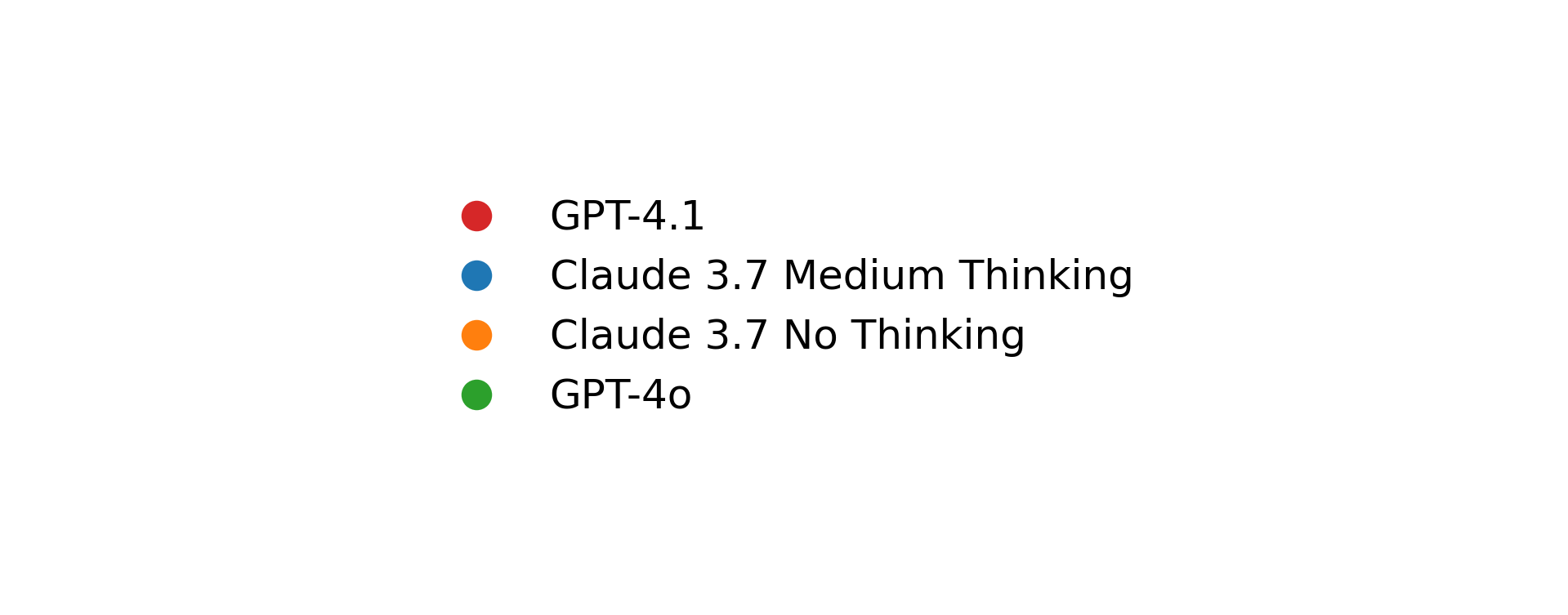}
    \caption{Color coding of data; all following plots use this format}
    \label{fig:legend}
\end{figure}

\FloatBarrier
\newpage

\subsection{Transformation}

\begin{table}[h!]
    \centering
    \begin{tabular}{|c|c|c|}
    \hline
        Models & Smallest 5 Grids & Largest 5 Grids \\
        \hline
        GPT-4o & 6\% & 0\% \\
        \hline
        GPT-4.1 & 14\% & 0\% \\
        \hline
        Claude No Thinking & 62\% & 0\% \\
        \hline
        Claude Medium Thinking & 70\% & 0\% \\
        \hline
    \end{tabular}
    \caption{Model Accuracy on Small and Large Grids (Transform)}
    \label{tab:transform}
\end{table}

\FloatBarrier

\begin{figure}[h!]
    \centering
    \begin{subfigure}{0.45\textwidth}
        \centering
        \includegraphics[width=\linewidth]{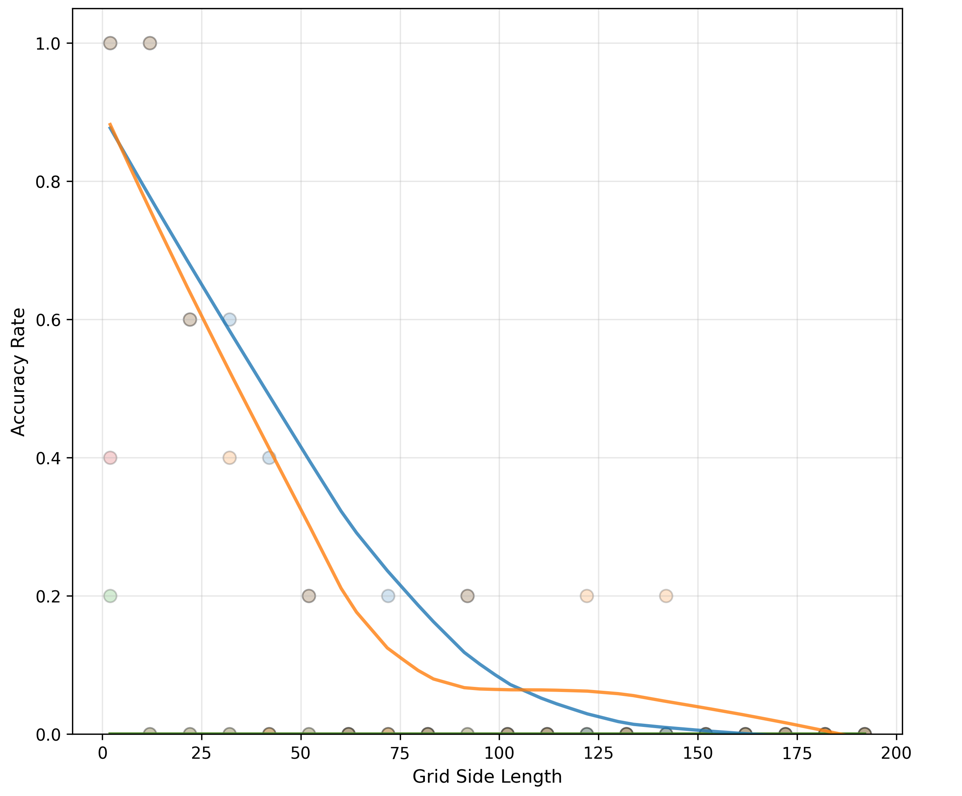}
        \caption{Transform over X-Axis}
        \label{fig:transform_x}
    \end{subfigure}
    \centering
    \begin{subfigure}{0.45\textwidth}
        \centering
        \includegraphics[width=\linewidth]{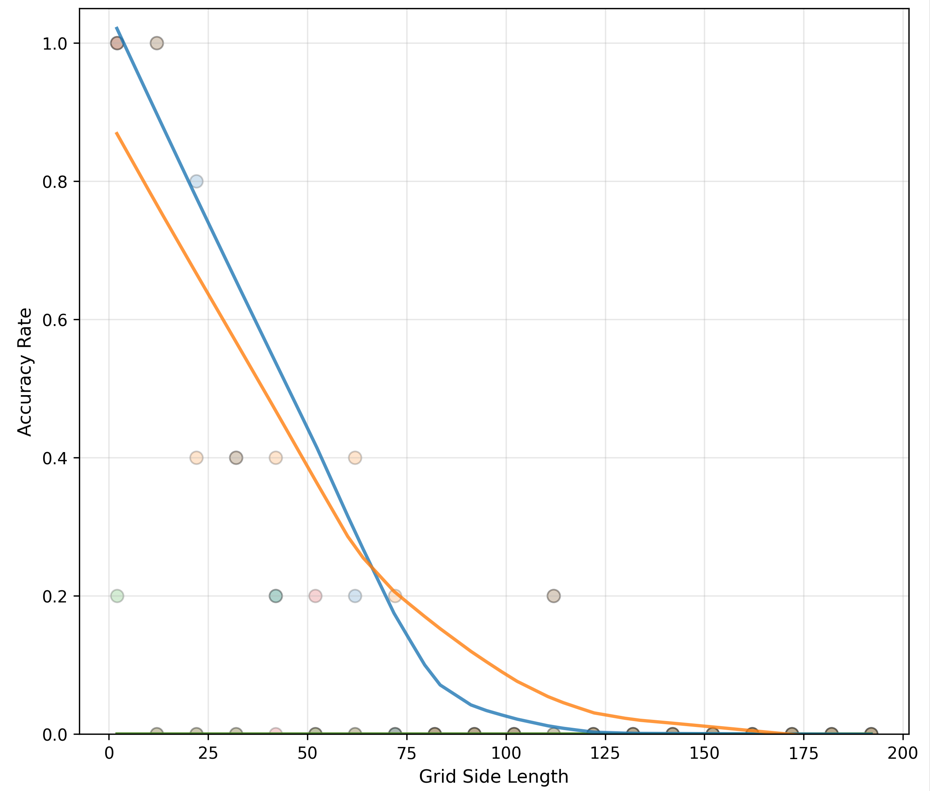}
        \caption{Transform over Y-Axis}
        \label{fig:transform_y}
    \end{subfigure}
    \caption{Accuracy vs Grid Size (Transformation)}
    \label{fig:trans_accuracy}
\end{figure}

Our suspicions about errors from the Quadrant test stemming from the models misreading the initial board were quickly proven in the Transformation test. Grid sizes ranged from 2$\times$2 to 200$\times$200, with each grid testing reflections across both axes. The models' inability to properly read board states was immediately evident: even on small boards, the OpenAI models averaged under 50\% accuracy. Performance deteriorated much quicker as well, with Anthropic models dropping below 20\% accuracy by the 75$\times$75 grid and the OpenAI models degrading even faster. By the largest grids, models rarely, if ever, got the answer correct. These trends can be seen in Figure~\ref{fig:trans_accuracy}.

\begin{figure}[h!]
    \centering
    \includegraphics[width=0.6\linewidth]{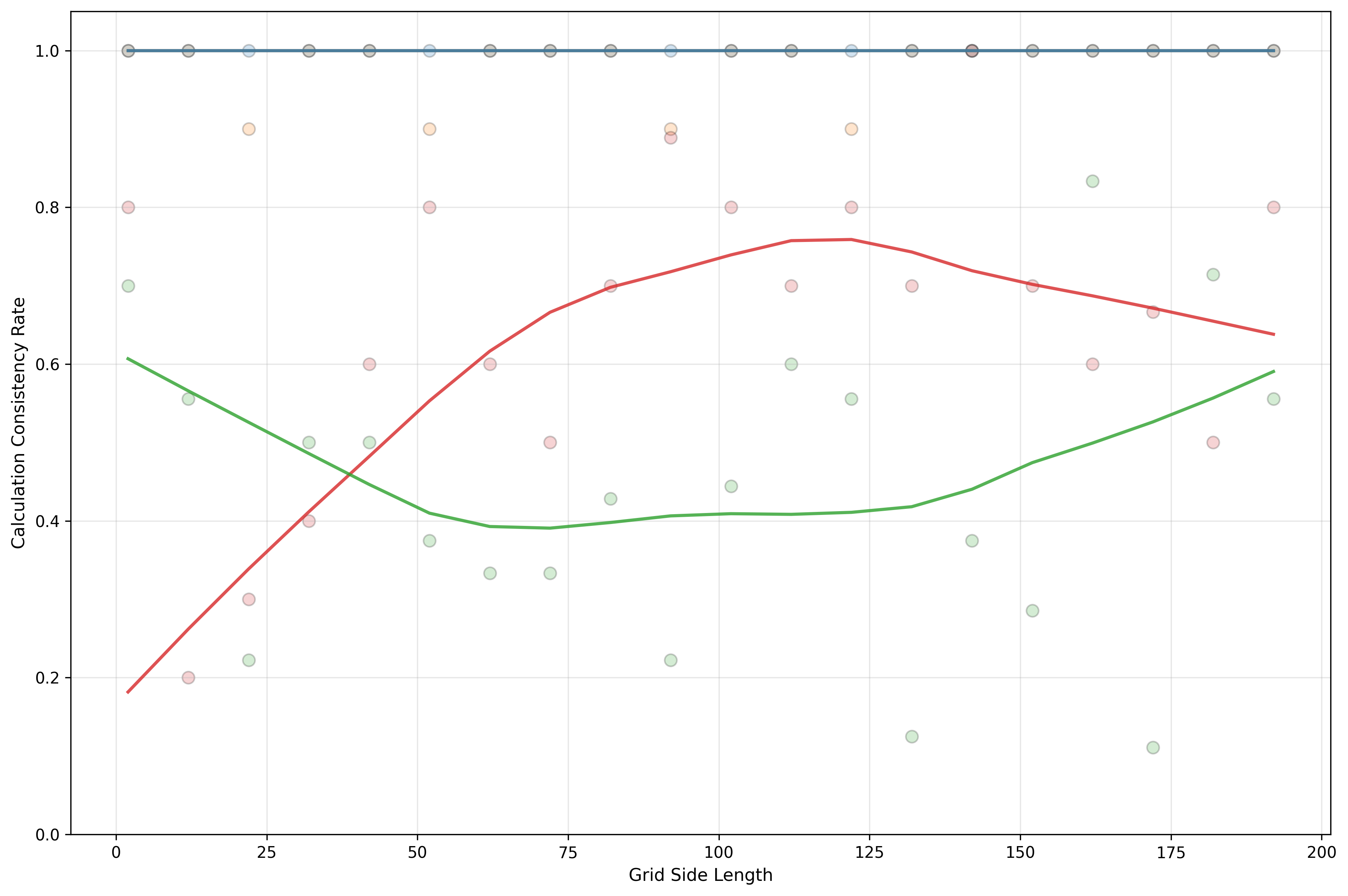}
    \caption{Consistency vs Grid Size (Transformation)}
    \label{fig:trans_consistency}
\end{figure}

It seems that the margin of error built into the nature of the Quadrant test shielded the models from such immediate deterioration. Interestingly, the models were not as internally consistent in this test as well. Both Anthropic models performed very well, with perfect internal consistency whereas the OpenAI models sometimes hit as low as 20\% calculation accuracy for certain grids. There appears to be no definite pattern for which grids were inconsistent, and the results can be seen in Figure~\ref{fig:trans_consistency}.

\FloatBarrier
\newpage

\subsection{Distance}

\begin{table}[h!]
    \centering
    \begin{tabular}{|c|c|c|}
    \hline
        Models & Smallest 5 Grids & Largest 5 Grids \\
        \hline
        GPT-4o & 44\% & 0\% \\
        \hline
        GPT-4.1 & 76\% & 8\% \\
        \hline
        Claude No Thinking & 96\% & 40\% \\
        \hline
        Claude Medium Thinking & 100\% & 16\% \\
        \hline
    \end{tabular}
    \caption{Model Accuracy on Small and Large Grids (Distance)}
    \label{tab:distance}
\end{table}

\FloatBarrier

\begin{figure}[h!]
    \centering
    \begin{subfigure}{0.45\textwidth}
        \centering
        \includegraphics[width=\linewidth]{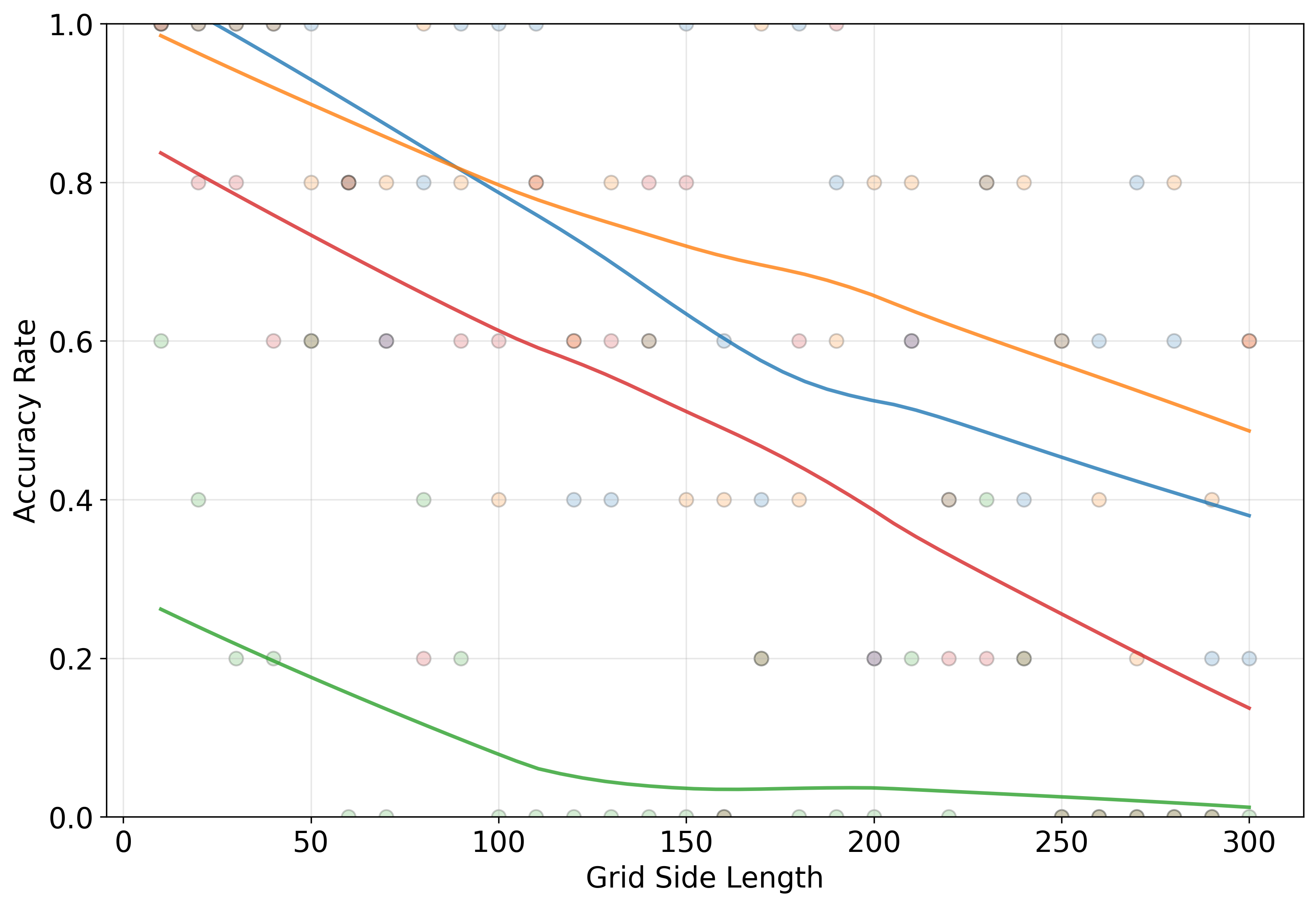}
        \caption{Closest Letter}
        \label{fig:distance_closest}
    \end{subfigure}
    \centering
    \begin{subfigure}{0.45\textwidth}
        \centering
        \includegraphics[width=\linewidth]{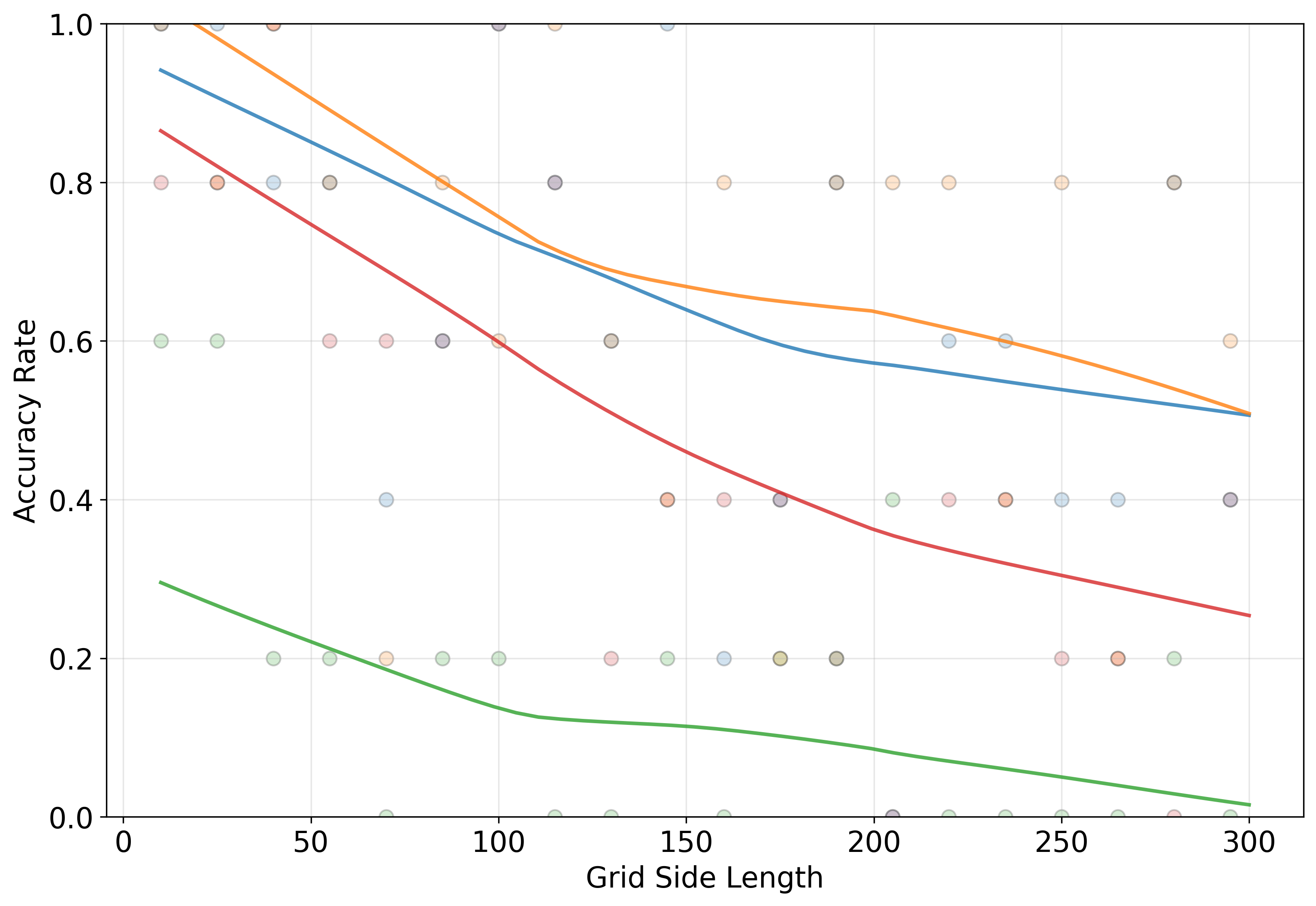}
        \caption{Farthest Letter}
        \label{fig:distance_farthest}
    \end{subfigure}
    \caption{Accuracy vs Grid Size (Distance)}
    \label{fig:dist_accuracy}
\end{figure}

Given that the failures of the previous two tests were largely due to miscounting, we devised this test to evaluate a model's ability to reason relatively; that is, in relation to other objects within a geometric space. Models were tested on grids ranging from 10$\times$10 to 200$\times$200, with a step of 5. Like the previous tests, models were more accurate on smaller boards and their performance degraded as grid size increased, as shown in Figure~\ref{fig:dist_accuracy}. 

\begin{figure}[h!]
    \centering
    \includegraphics[width=0.9\linewidth]{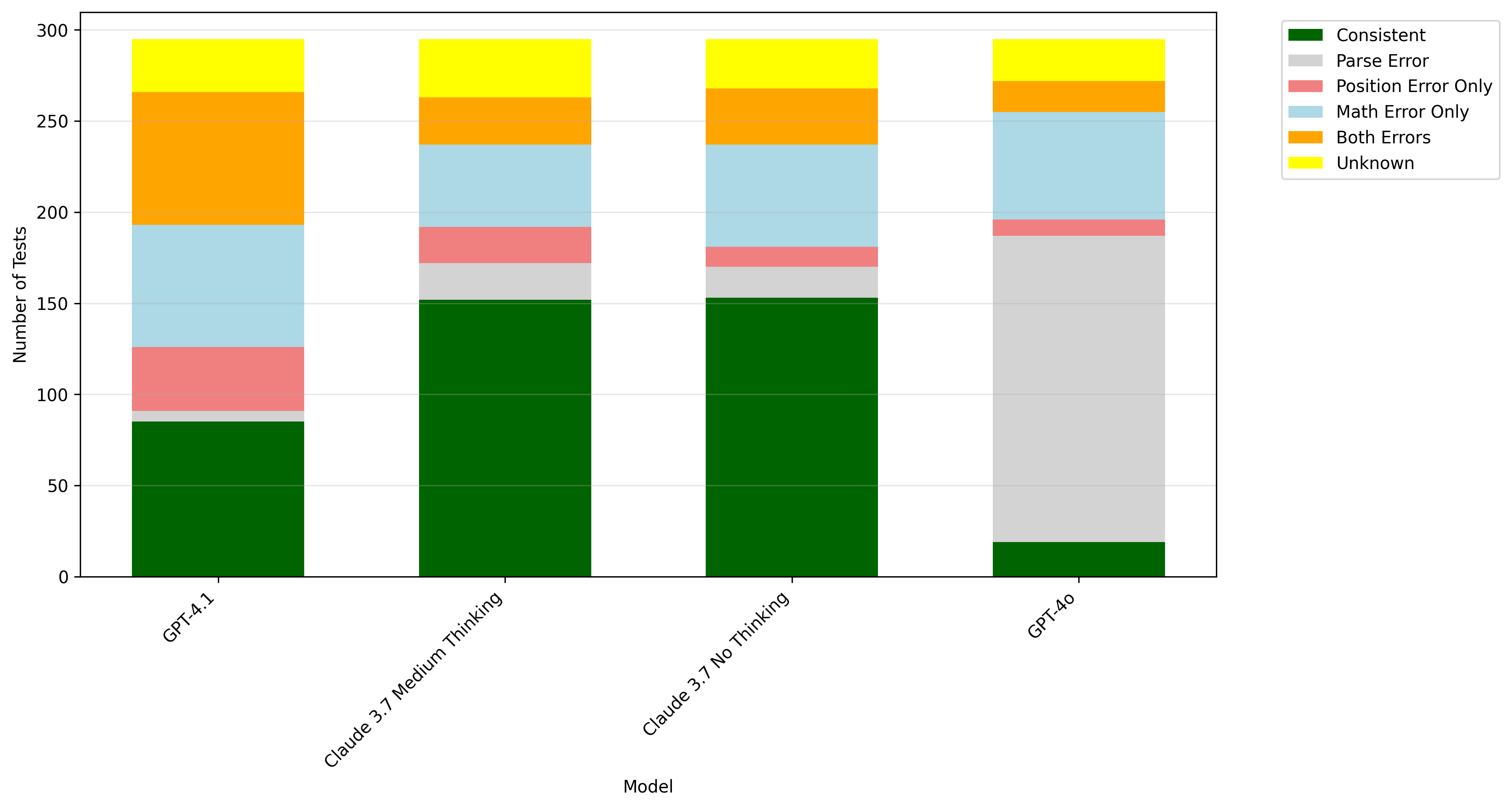}
    \caption{Error Types (Distance)}
    \label{fig:dist_consistency}
\end{figure}

Interestingly, models did not exhibit much internal consistency, especially compared to the previous tests. They frequently made mathematical errors, accounting for around half of the errors for each model, as seen in Figure~\ref{fig:dist_consistency}.

\begin{figure}[h!]
    \centering
    \includegraphics[width=0.75\linewidth]{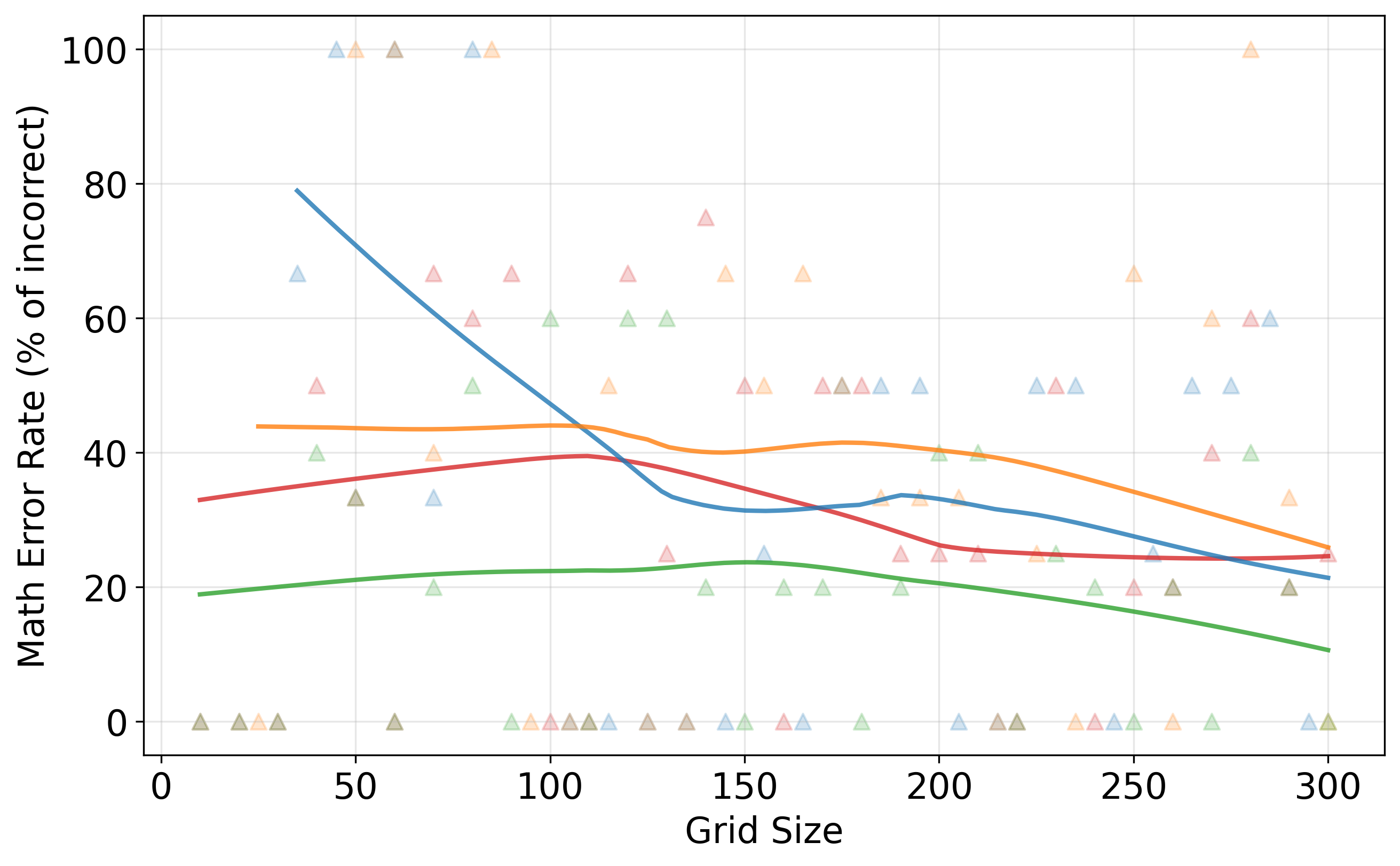}
    \caption{Math Error vs Grid Size (Distance)}
    \label{fig:dist_math}
\end{figure}

The math error rate decreased as grid size increased for the OpenAI models and remained roughly constant for the Anthropic models. The positional error rate increased with grid size, similar to other tests. The results are shown in Figure~\ref{fig:dist_math}.

\FloatBarrier
\newpage

\subsection{Word Search}

\begin{table}[h!]
    \centering
    \begin{tabular}{|c|c|c|}
    \hline
        Models & Smallest 5 Grids & Largest 5 Grids \\
        \hline
        \rowcolor{gray!15}
        GPT-4o; 3 Letters& 15\% & 0\% \\
        \hline
        \rowcolor{gray!15}
        GPT-4.1; 3 Letters & 72.5\% & 0\% \\
        \hline
        \rowcolor{gray!15}
        Claude No Thinking; 3 Letters & 95\% & 2\% \\
        \hline
        \rowcolor{gray!15}
        Claude Medium Thinking; 3 Letters & 95\% & 0\% \\
        \hline \hline
        GPT-4o; 6 Letters& 0\% & 0\% \\
        \hline
        GPT-4.1; 6 Letters & 20\% & 0\% \\
        \hline
        Claude No Thinking; 6 Letters & 30\% & 0\% \\
        \hline
        Claude Medium Thinking; 6 Letters & 35\% & 0\% \\
        \hline \hline
        \rowcolor{gray!15}
        GPT-4o; 9 Letters& 5.5\% & 0\% \\
        \hline
        \rowcolor{gray!15}
        GPT-4.1; 9 Letters & 47.2\% & 0\% \\
        \hline
        \rowcolor{gray!15}
        Claude No Thinking; 9 Letters & 80.5\% & 0\% \\
        \hline
        \rowcolor{gray!15}
        Claude Medium Thinking; 9 Letters & 72.2\% & 0\% \\
        \hline
    \end{tabular}
    \caption{Model Accuracy on Small and Large Grids (Search)}
    \label{tab:search}
\end{table}

\begin{figure}[h!]
    \centering
    \begin{subfigure}{0.45\textwidth}
        \centering
        \includegraphics[width=\linewidth]{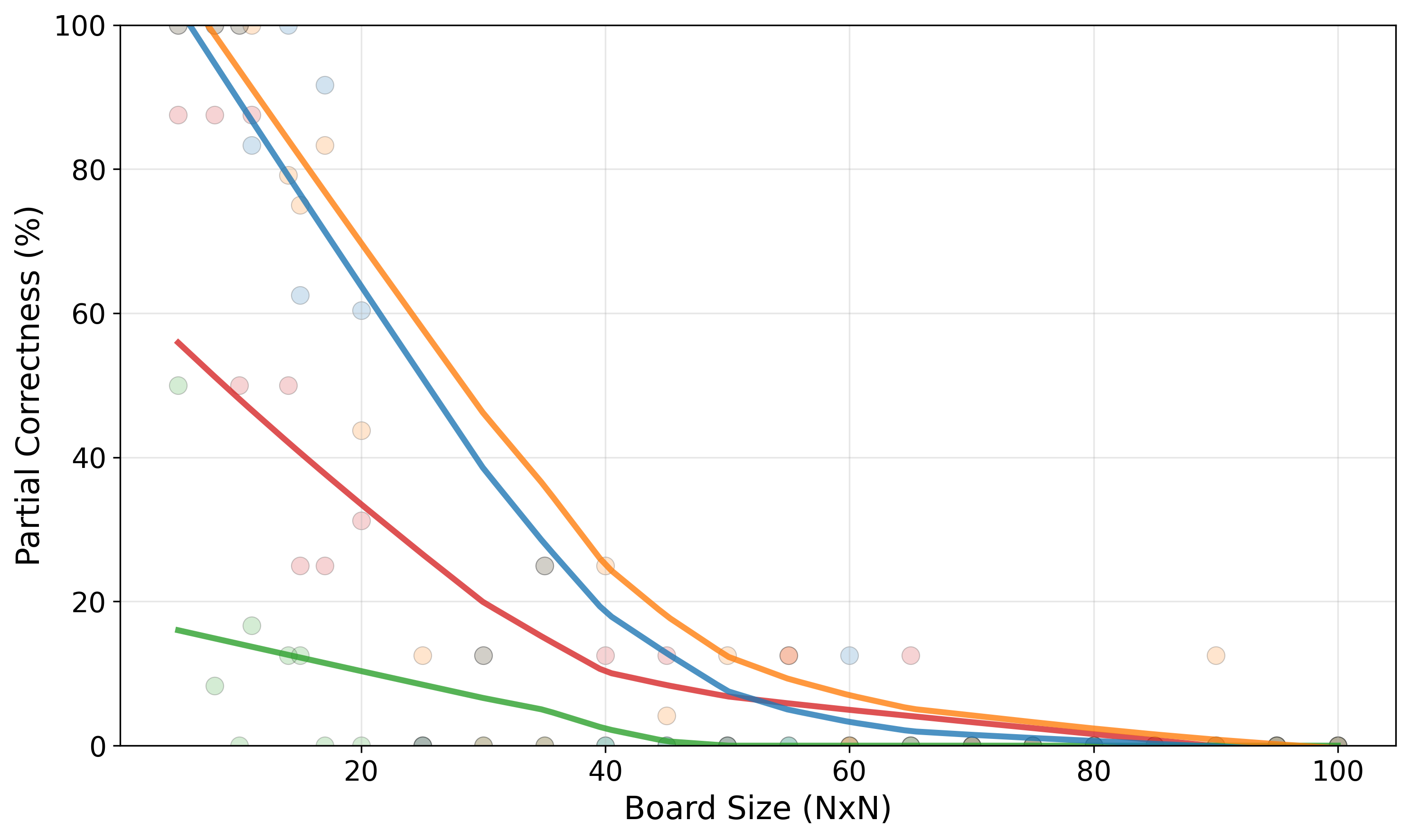}
        \caption{3 Letter Words}
        \label{fig:search_accuracy_1}
    \end{subfigure}
    \centering
    \begin{subfigure}{0.45\textwidth}
        \centering
        \includegraphics[width=\linewidth]{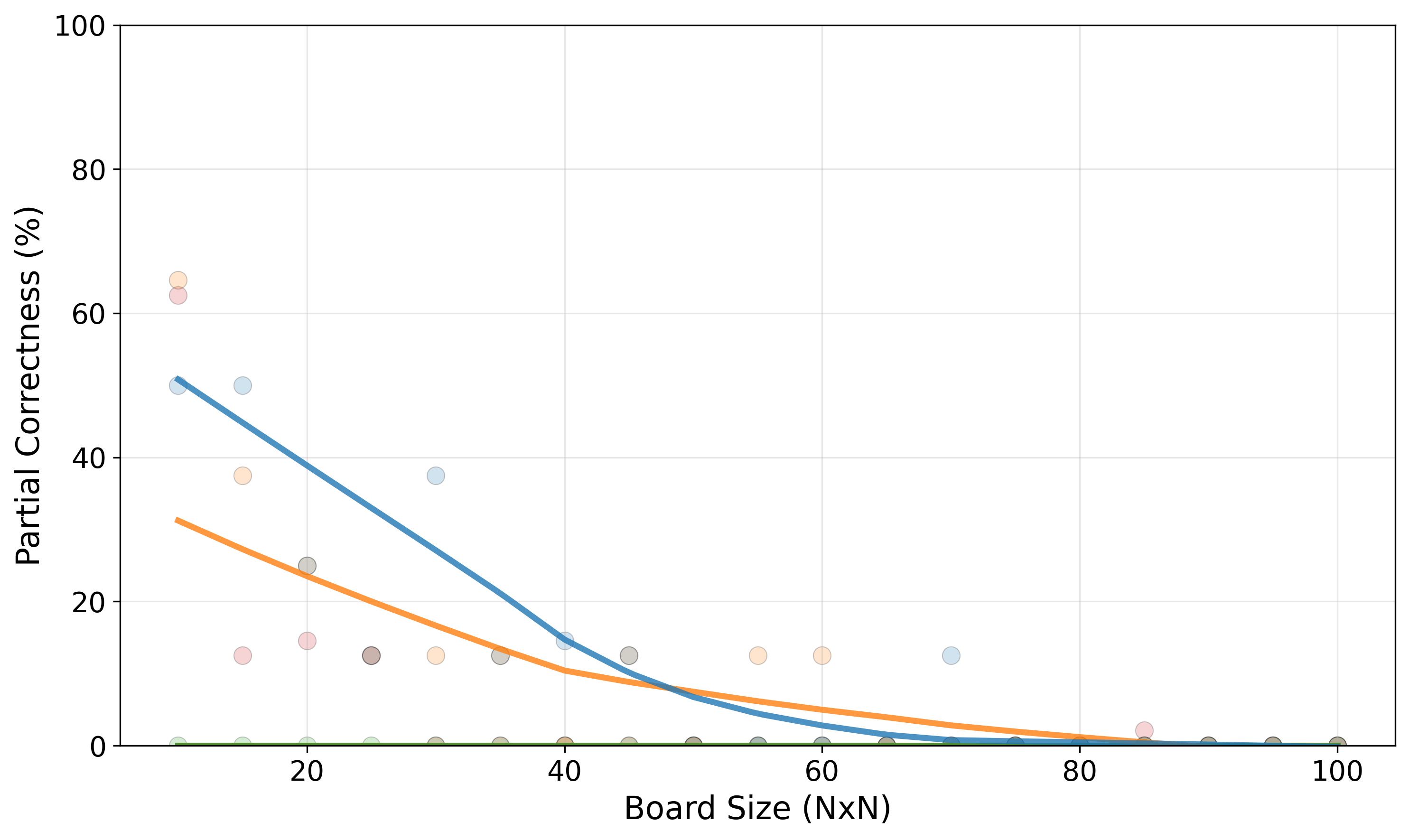}
        \caption{6 Letter Words}
        \label{fig:search_accuracy_2}
    \end{subfigure}
    \centering
    \begin{subfigure}{0.45\textwidth}
        \centering
        \includegraphics[width=\linewidth]{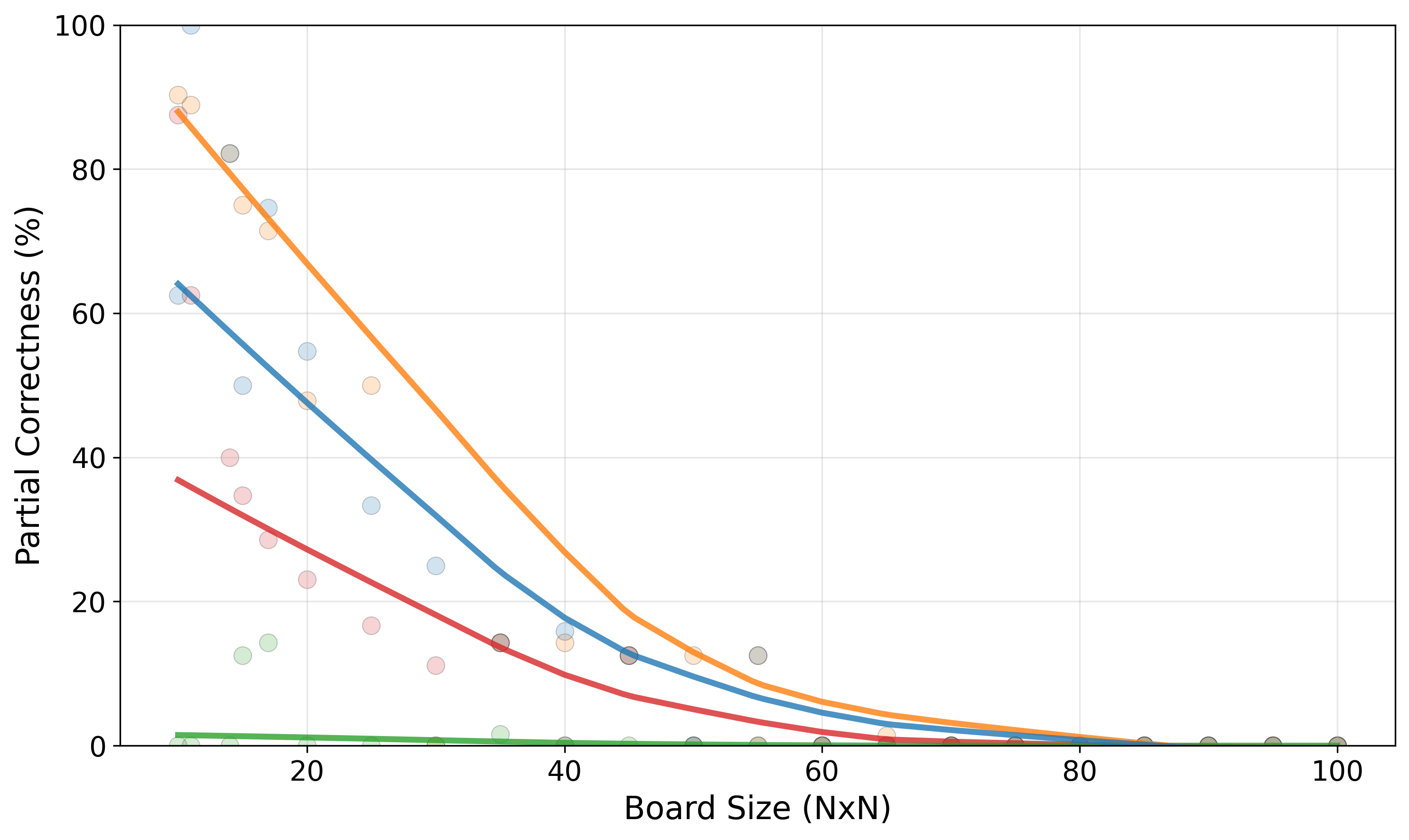}
        \caption{9 Letter Words}
        \label{fig:search_accuracy_3}
    \end{subfigure}
    \caption{Accuracy vs Grid Size (Search)}
    \label{fig:search_accuracy}
\end{figure}

This test was non-numerical in nature. Instead of calculations, we tested the models on language -- a subject in which they traditionally excel. Grid sizes were tested from 5$\times$5 to 100$\times$100 with a step of 5, using words of length 3, 6, and 9. Words that did not fit in the board, ie. a nine-letter word for a 5$\times$5 grid, were omitted from testing. All eight spelling directions (horizontal right-to-left, vertical up-to-down, etc) were tested for each word and grid size. Credit was given for percentage of coordinates correct, ranging from 0 to 1. This test was more difficult for the models from the start. The Anthropic models, which averaged close to 100\% accuracy on smaller boards in the other tests, averaged only 50\% for the six-letter words. The results for all three word lengths are seen in Figure~\ref{fig:search_accuracy}.

\begin{figure}[h!]
    \centering
    \begin{subfigure}{0.45\textwidth}
        \centering
        \includegraphics[width=\linewidth]{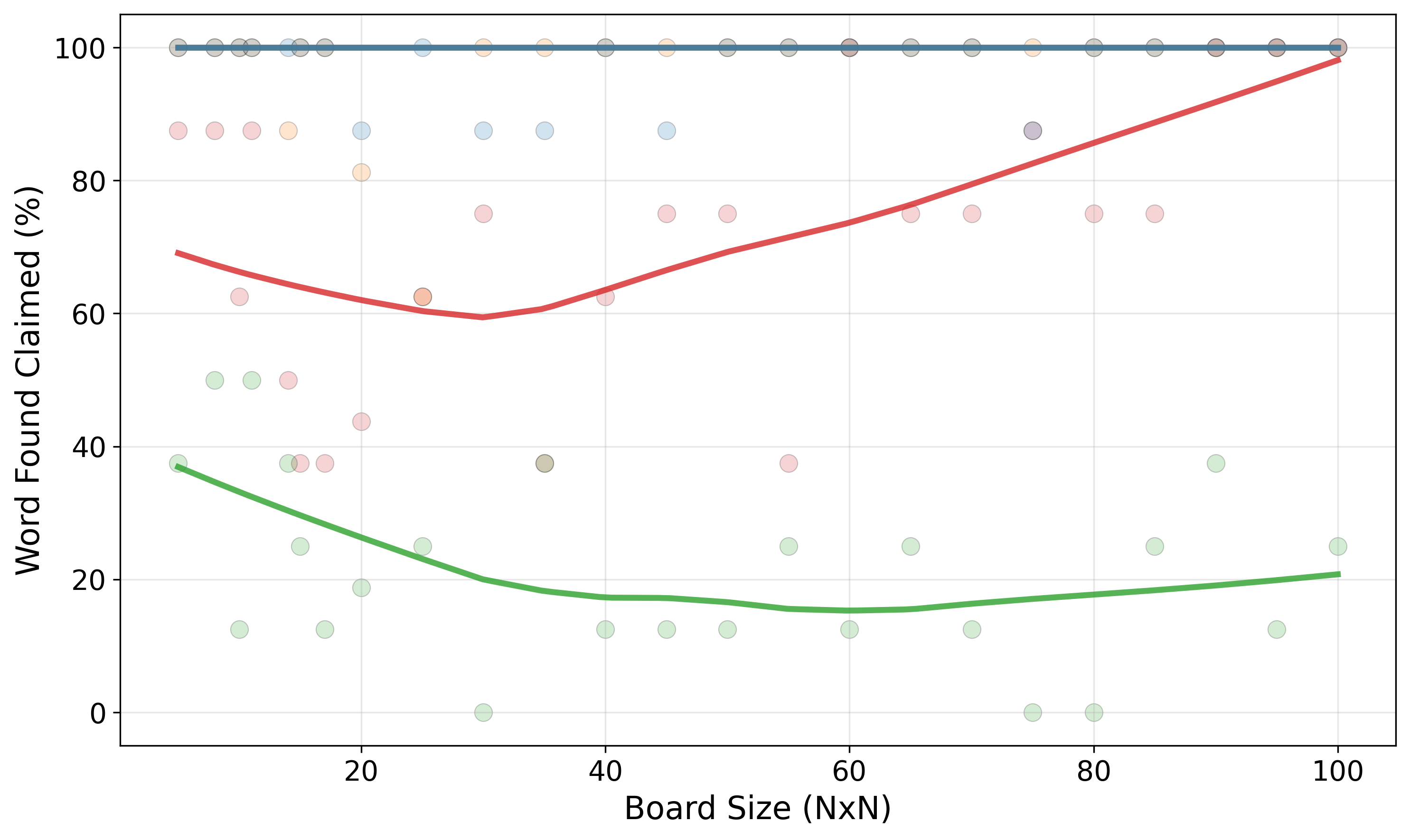}
        \caption{3 Letter Words}
        \label{fig:search_claim_1}
    \end{subfigure}
    \centering
    \begin{subfigure}{0.45\textwidth}
        \centering
        \includegraphics[width=\linewidth]{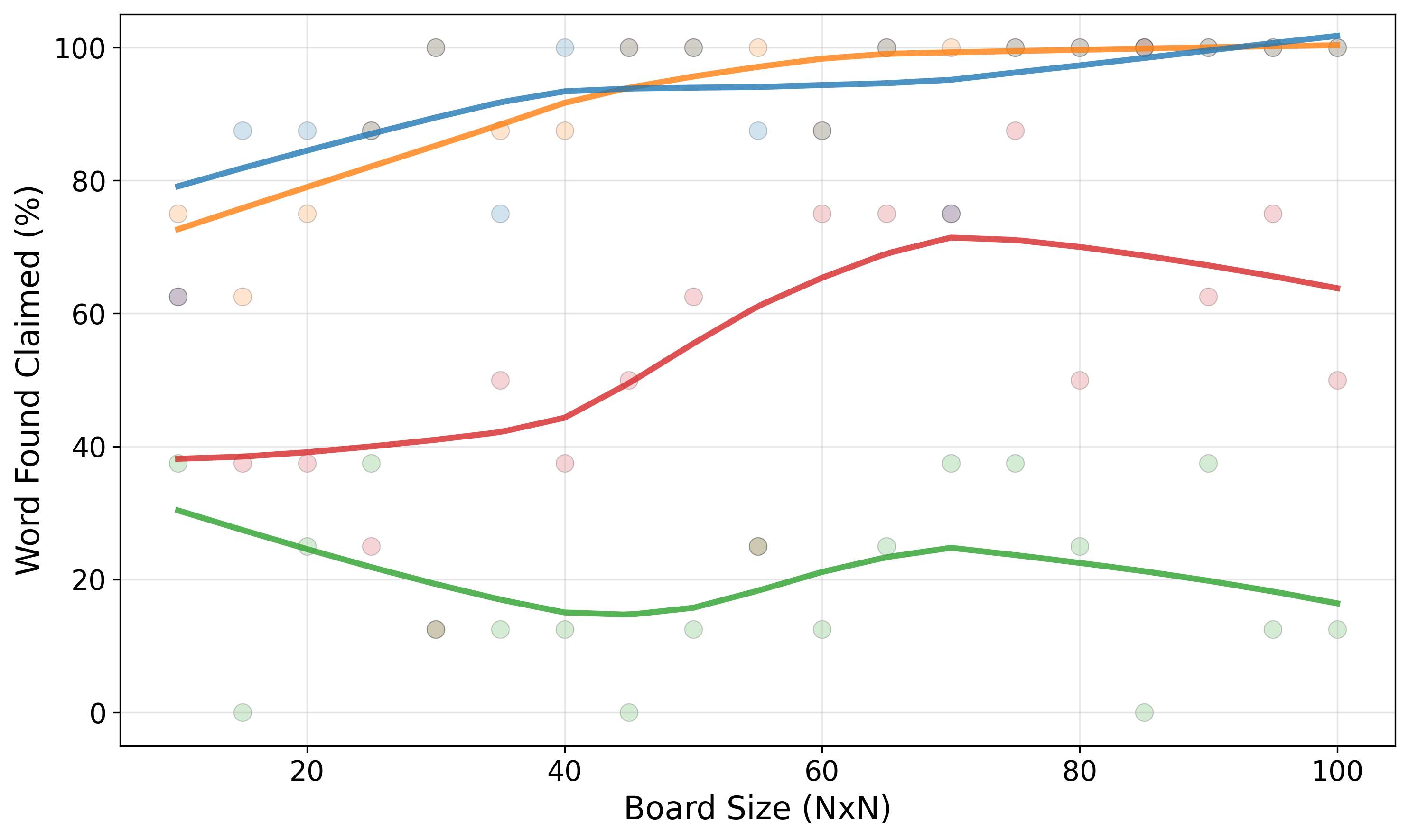}
        \caption{6 Letter Words}
        \label{fig:search_claim_2}
    \end{subfigure}
    \centering
    \begin{subfigure}{0.45\textwidth}
        \centering
        \includegraphics[width=\linewidth]{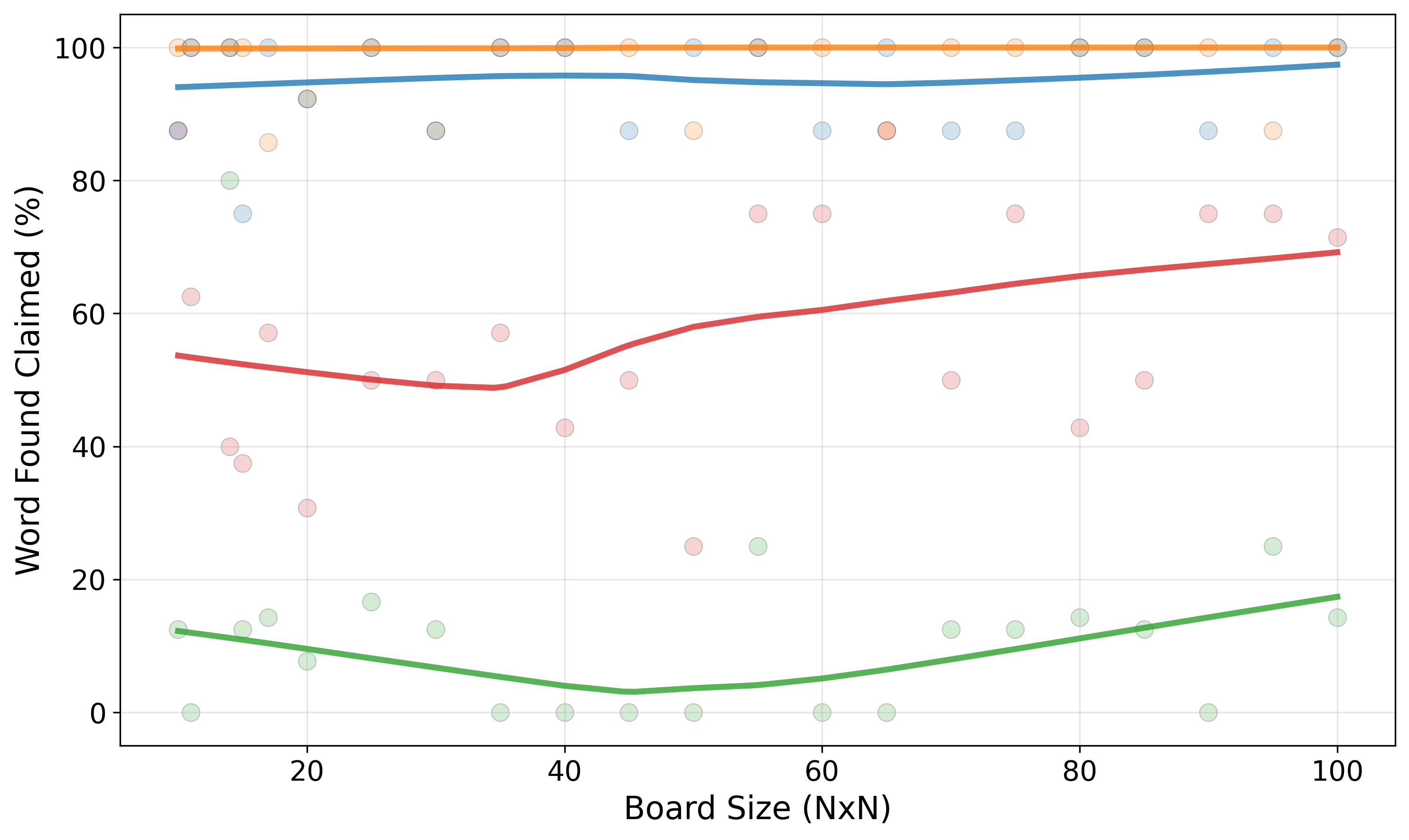}
        \caption{9 Letter Words}
        \label{fig:search_claim_3}
    \end{subfigure}
    \caption{Claimed Detection vs Grid Size (Search)}
    \label{fig:search_claims}
\end{figure}

Despite the low accuracy, however, the models still often claimed to have found the word. The Anthropic models had an almost 100\% record of claimed detection, and the OpenAI models were somewhat lower. This was likely due to hallucination, the effects of which can be seen in Figure~\ref{fig:search_claims}.

\begin{figure}[h!]
    \centering
    \includegraphics[width=1\linewidth]{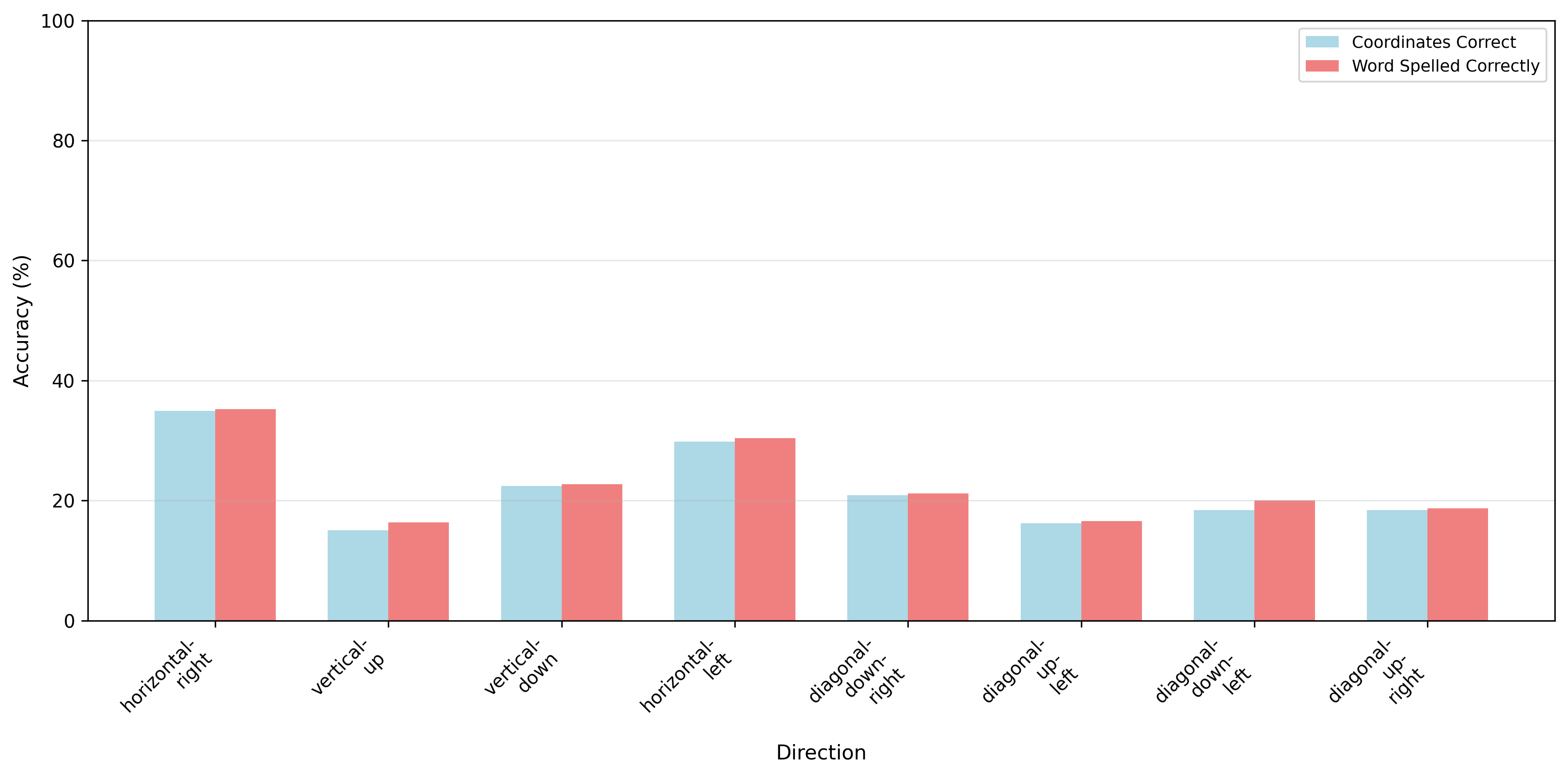}
    \caption{Accuracy by Direction (Search)}
    \label{fig:search_direction}
\end{figure}

Additionally, models seemed to have trouble identifying words across multiple columns and in unusual orientations. The average accuracy rate was highest for words spelled traditionally, horizontally left-to-right, second-highest backwards spelling, horizontal right-to-left, and lowest for vertical spelling, down-to-up, as shown in Figure~\ref{fig:search_direction}.

\FloatBarrier
\newpage
\newpage
\subsection{Slide}

\begin{figure}[h!]
    \centering
    \includegraphics[width=0.9\linewidth]{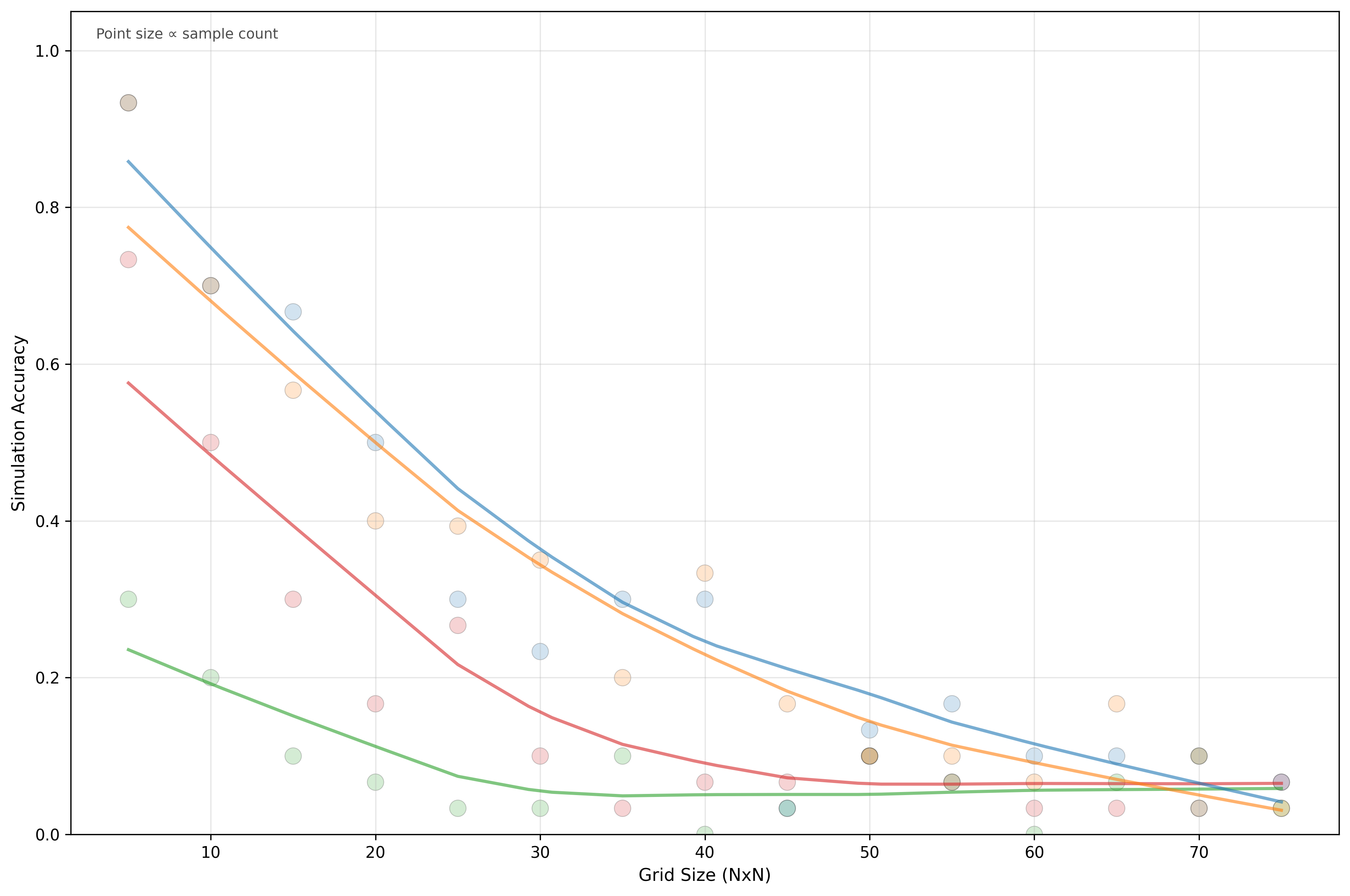}
    \caption{Accuracy vs Grid Size (Slide)}
    \label{fig:slide_accuracy}
\end{figure}

\begin{table}[h!]
    \centering
    \begin{tabular}{|c|c|c|}
    \hline
        Models & Smallest 5 Grids & Largest 5 Grids \\
        \hline
        \rowcolor{gray!15}
        GPT-4o; 1 slide & 28\% & 8\% \\
        \hline
        \rowcolor{gray!15}
        GPT-4.1; 1 slide & 64\% & 16\% \\
        \hline
        \rowcolor{gray!15}
        Claude No Thinking; 1 slide & 92\% & 2\% \\
        \hline
        \rowcolor{gray!15}
        Claude Medium Thinking; 1 slide & 88\% & 20\% \\
        \hline
        \hline
        GPT-4o; 2 slides & 28\% & 16\% \\
        \hline
        GPT-4.1; 2 slides & 32\% & 4\% \\
        \hline
        Claude No Thinking; 2 slides & 76\% & 8\% \\
        \hline
        Claude Medium Thinking; 2 slides & 88\% & 24\% \\
        \hline
        \hline
        \rowcolor{gray!15}
        GPT-4o; 3 slides & 12\% & 0\% \\
        \hline
        \rowcolor{gray!15}
        GPT-4.1; 3 slides & 44\% & 0\% \\
        \hline
        \rowcolor{gray!15}
        Claude No Thinking; 3 slides & 64\% & 12\% \\
        \hline
        \rowcolor{gray!15}
        Claude Medium Thinking; 3 slides & 64\% & 4\% \\
        \hline
        \hline
        GPT-4o; 4 slides & 0\% & 8\% \\
        \hline
        GPT-4.1; 4 slides & 40\% & 12\% \\
        \hline
        Claude No Thinking; 4 slides & 56\% & 4\% \\
        \hline
        Claude Medium Thinking; 4 slides & 56\% & 4\% \\
        \hline
        \hline
        \rowcolor{gray!15}
        GPT-4o; 5 slides & 0\% & 0\% \\
        \hline
        \rowcolor{gray!15}
        GPT-4.1; 5 slides & 28\% & 4\% \\
        \hline
        \rowcolor{gray!15}
        Claude No Thinking; 5 slides & 36\% & 4\% \\
        \hline
        \rowcolor{gray!15}
        Claude Medium Thinking; 5 slides & 44\% & 0\% \\
        \hline
        \hline
        GPT-4o; 6 slides & 16\% & 0\% \\
        \hline
        GPT-4.1; 6 slides & 28\% & 0\% \\
        \hline
        Claude No Thinking; 6 slides & 32\% & 0\% \\
        \hline
        Claude Medium Thinking; 6 slides & 32\% & 4\% \\
        \hline
    \end{tabular}
    \caption{Model Accuracy on Small and Large Grids (Slide)}
    \label{tab:slide}
\end{table}
\FloatBarrier
This final task was designed to test relative geometry and reasoning. Grid sizes were tested from 5$\times$5 to 75$\times$75, with increments of 5$\times$5. Each grid size was tested with 1-6 slides. The final test showed a similar pattern of strong performance on initial, smaller boards and quick performance deterioration as grid sizes grew. The Anthropic models consistently outperformed the OpenAI ones until a grid size of 70$\times$70, after which all accuracy was effectively $0$. This deterioration was even quicker than in Word Search, implying that this test was overall more difficult despite the models having a more consistent performance on the smaller boards. When plotting accuracy, each slide of the multi-slide was treated independently to remove cascading effects. The results can be seen in Figure~\ref{fig:slide_accuracy}.

\begin{figure}[h!]
    \centering
    \begin{subfigure}{0.3\textwidth}
        \centering
        \includegraphics[width=\linewidth]{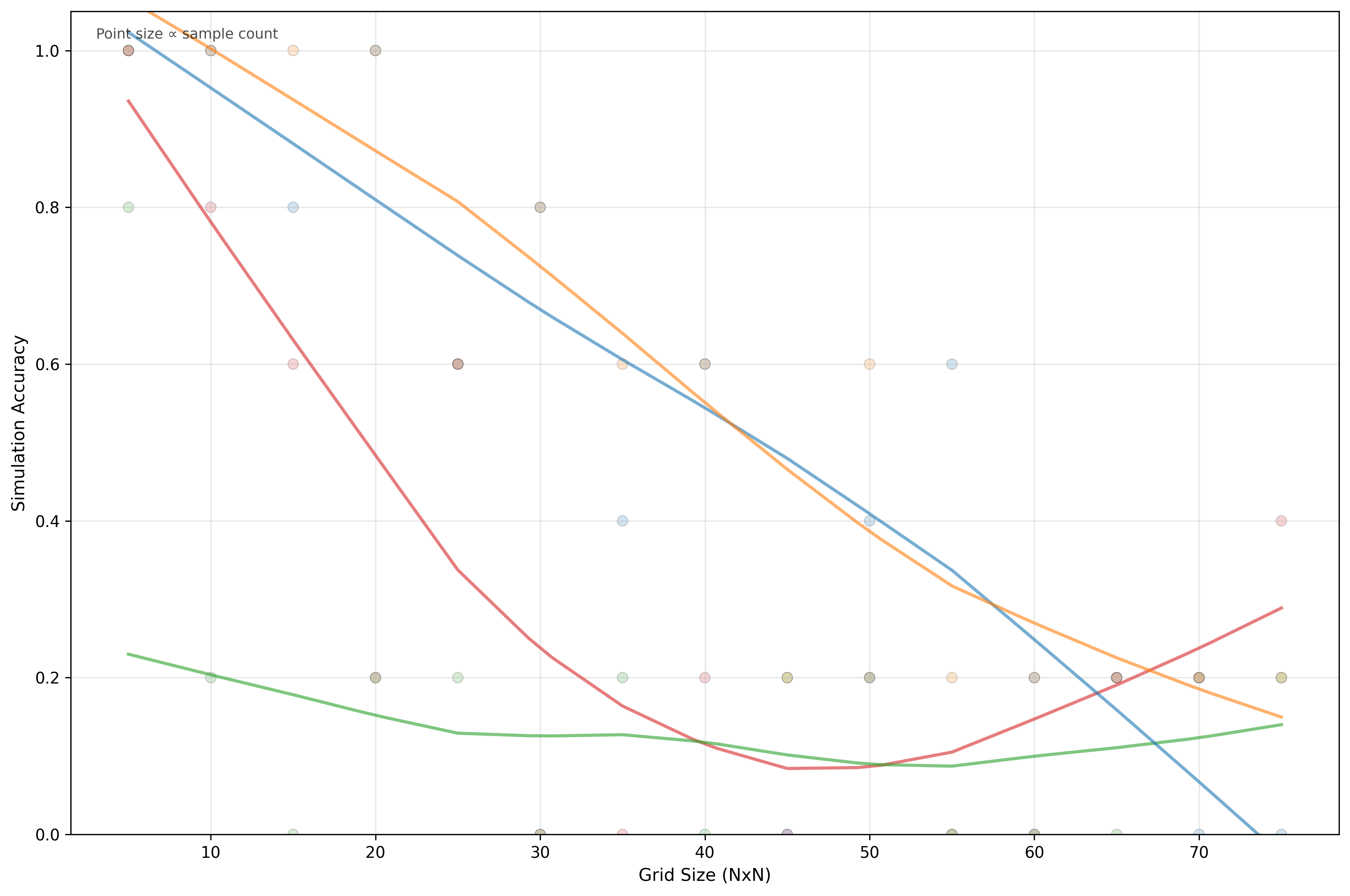}
        \caption{1 Slide}
        \label{fig:slide_1}
    \end{subfigure}
    \centering
    \begin{subfigure}{0.3\textwidth}
        \centering
        \includegraphics[width=\linewidth]{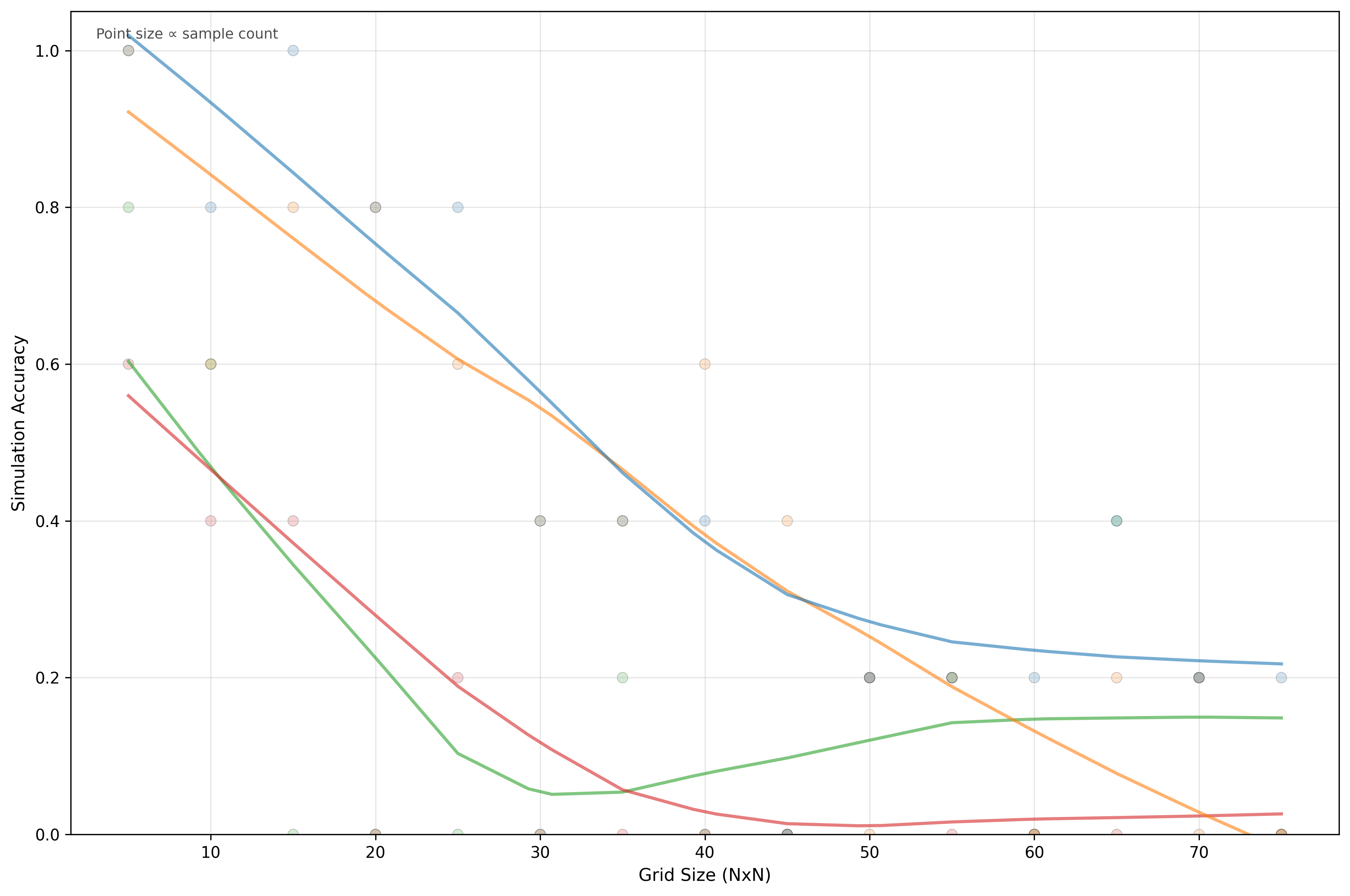}
        \caption{2 Slides}
        \label{fig:slide_2}
    \end{subfigure}
    \centering
    \begin{subfigure}{0.3\textwidth}
        \centering
        \includegraphics[width=\linewidth]{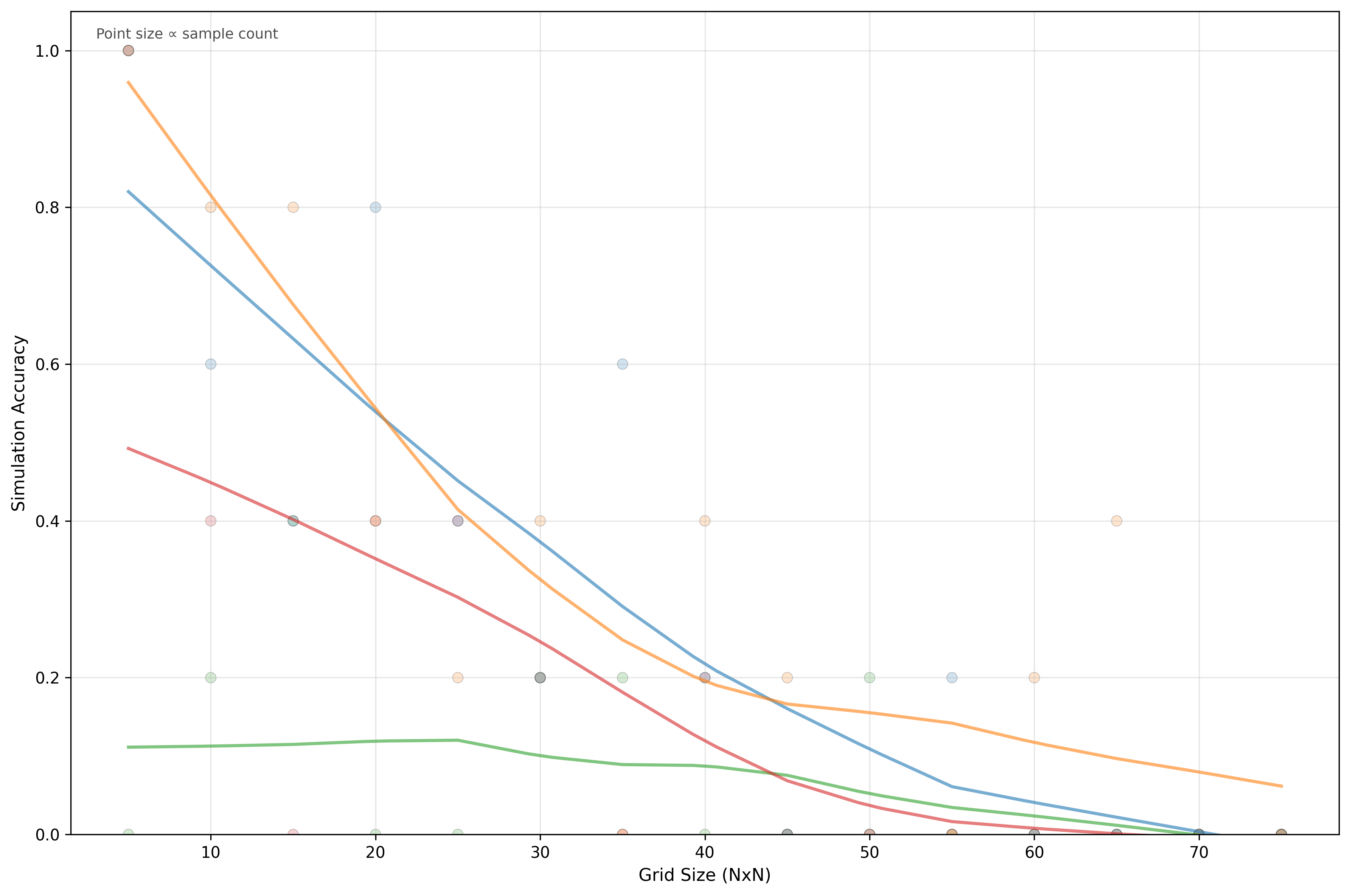}
        \caption{3 Slides}
        \label{fig:slide_3}
    \end{subfigure}
    \begin{subfigure}{0.3\textwidth}
        \centering
        \includegraphics[width=\linewidth]{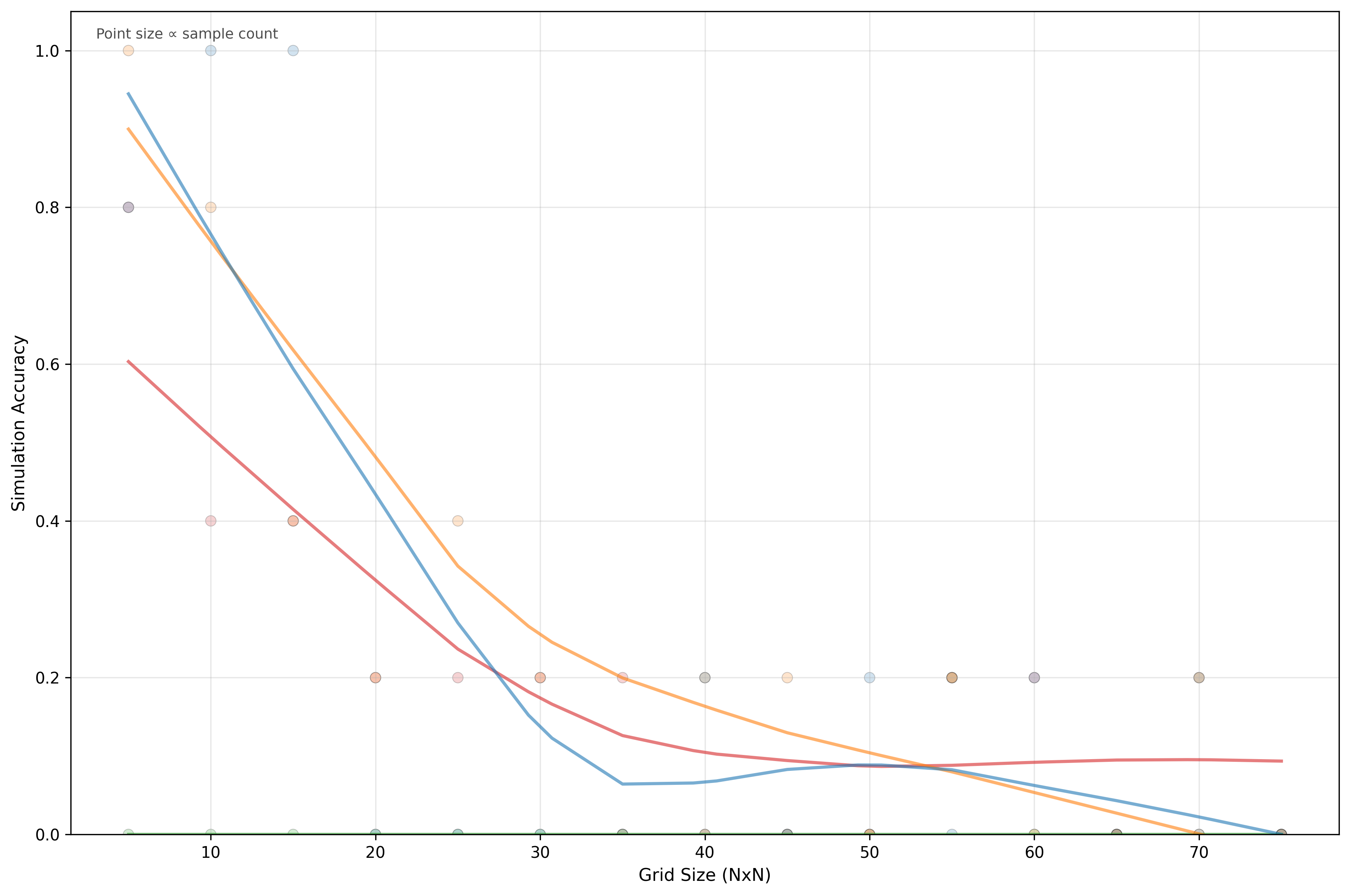}
        \caption{4 Slides}
        \label{fig:slide_4}
    \end{subfigure}
    \centering
    \begin{subfigure}{0.3\textwidth}
        \centering
        \includegraphics[width=\linewidth]{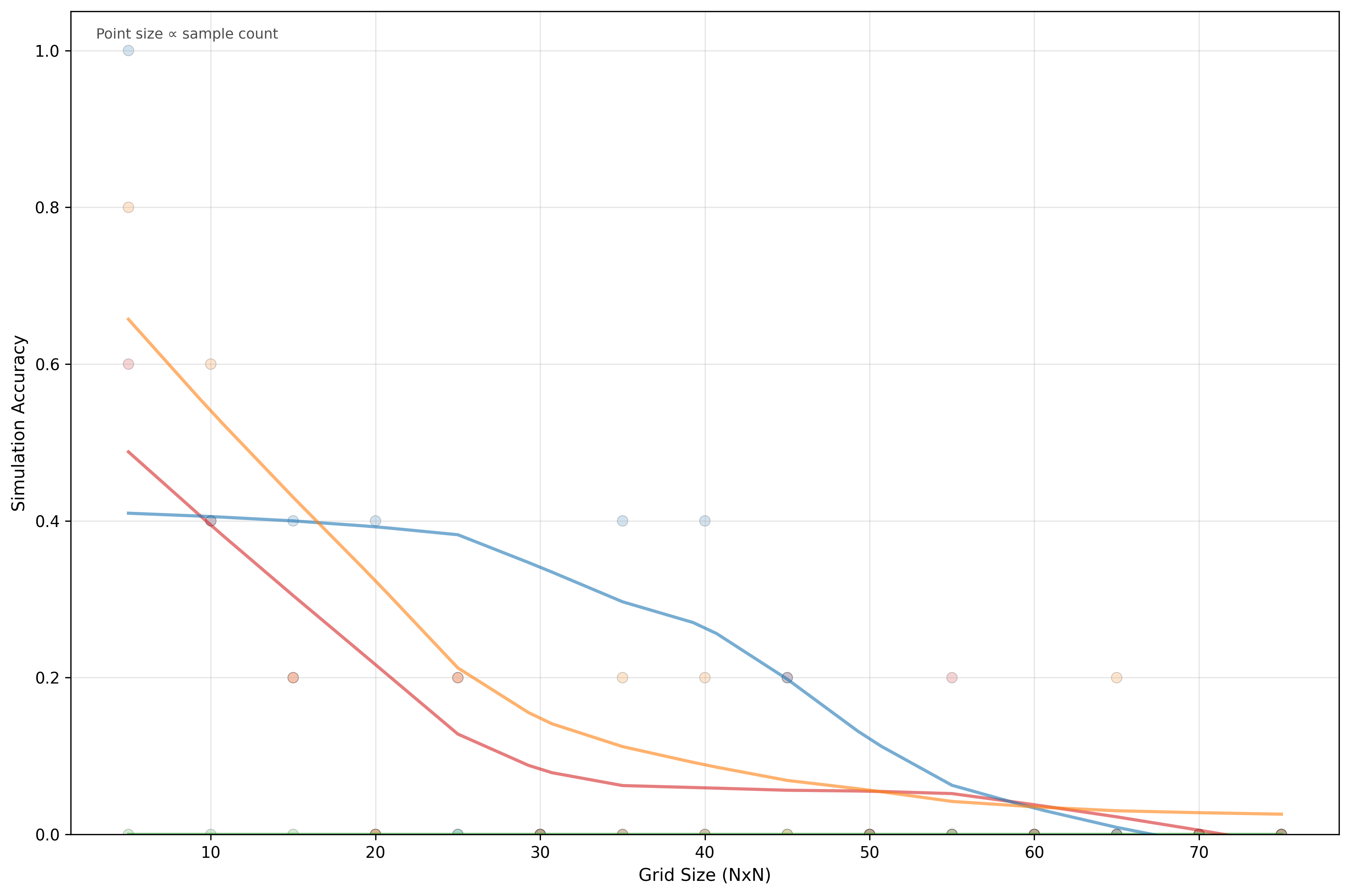}
        \caption{5 Slides}
        \label{fig:slide_5}
    \end{subfigure}
    \centering
    \begin{subfigure}{0.3\textwidth}
        \centering
        \includegraphics[width=\linewidth]{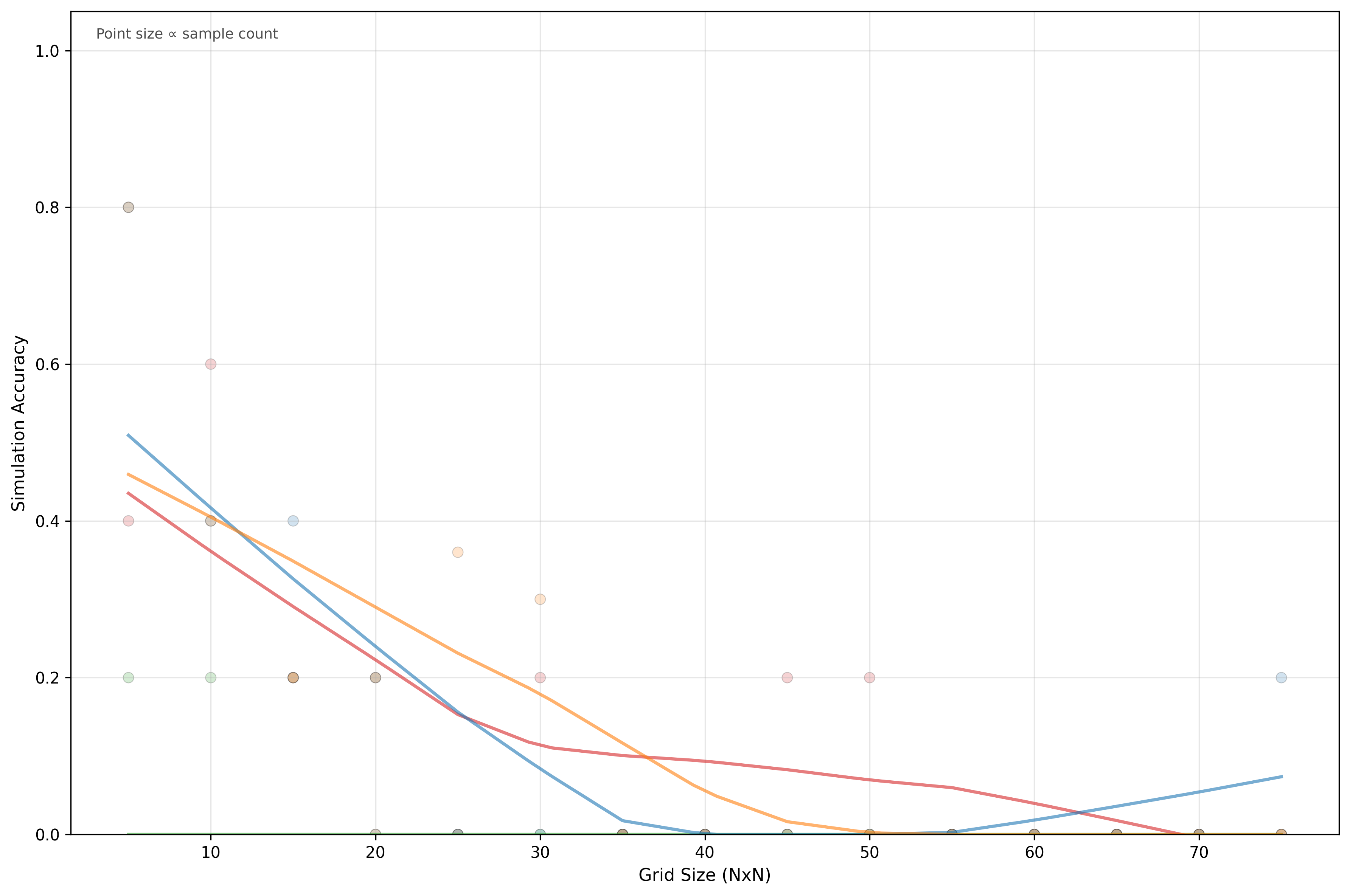}
        \caption{6 Slides}
        \label{fig:slide_6}
    \end{subfigure}
    \caption{Accuracy vs Grid Size per Slide Count (Slide)}
    \label{fig:slide_slide}
\end{figure}

The number of slides was also a determining factor in the models' average accuracy, with much quicker dropoffs in accuracy for prompts containing more slides. This was to be expected as the introduction of multi-step reasoning on top of the spatial reasoning likely compounded the difficulty. Even when asked to output intermediate steps the models showed no significant increase in performance, suggesting difficulty remembering where they had even placed the X before. A comparison of slides can be seen in Figure~\ref{fig:slide_slide}.

\begin{figure}[t]
    \centering
    \includegraphics[width=1\linewidth]{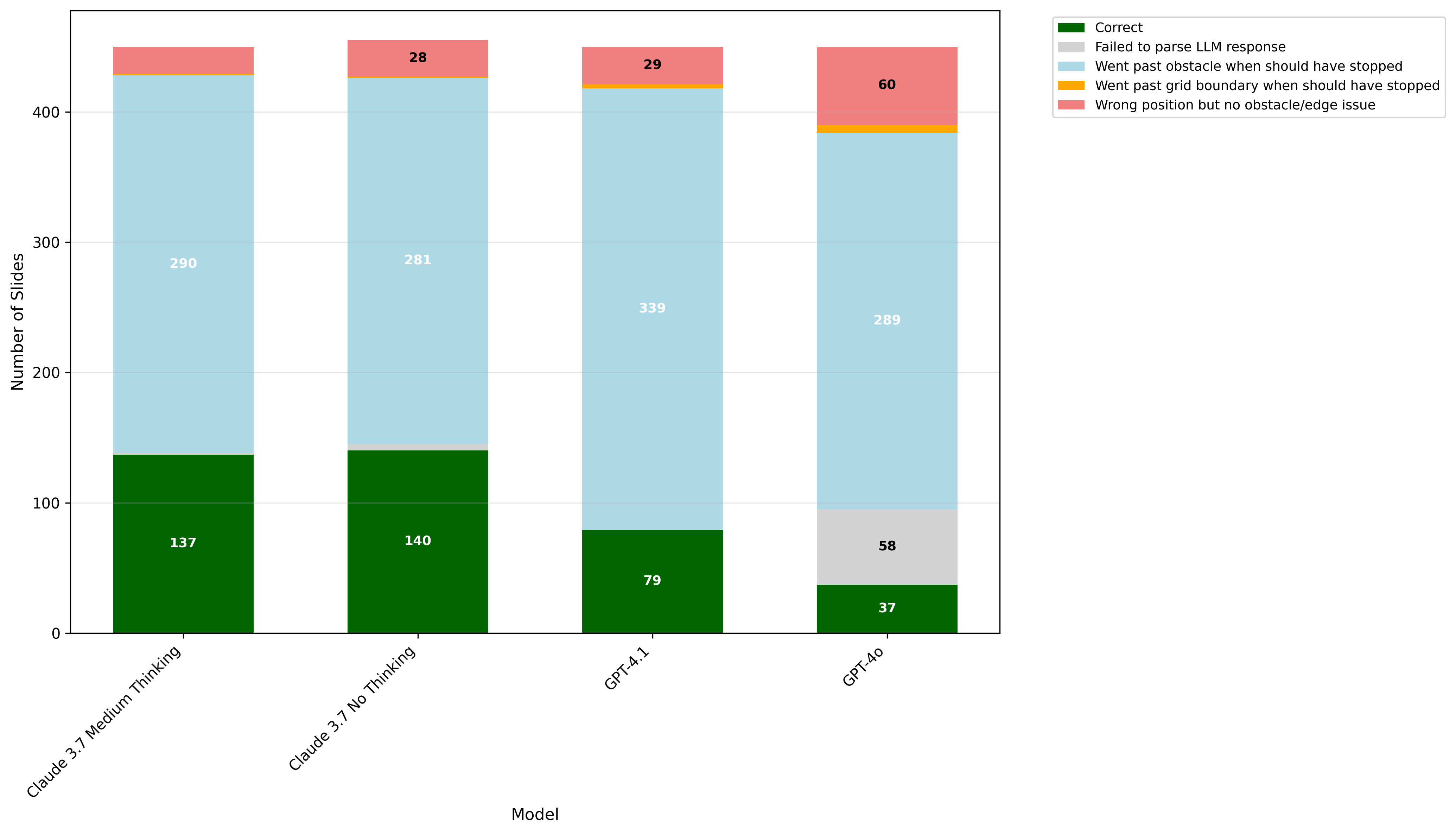}
    \caption{Error Types (Slide)}
    \label{fig:slide_error}
\end{figure}

The consistency is somewhat more difficult to parse for this test. The models often ignored walls but it remains unclear whether they miscounted the position of the walls while reading board states or experienced a systemic error with wall detection itself. There were still some classic miscounting issues as well, as shown in Figure~\ref{fig:slide_error}.

% \begin{figure}[h!]
%     \centering
%     \includegraphics[width=0.9\linewidth]{images/slide_all.png}
%     \caption{Accuracy vs Slide Count (Slide)}
%     \label{fig:slide_slide}
% \end{figure}

\clearpage

\FloatBarrier
\nopagebreak[4]
\subsection{Parsing}
\begin{figure}[h!]
    \centering
    \includegraphics[width=0.75\linewidth]{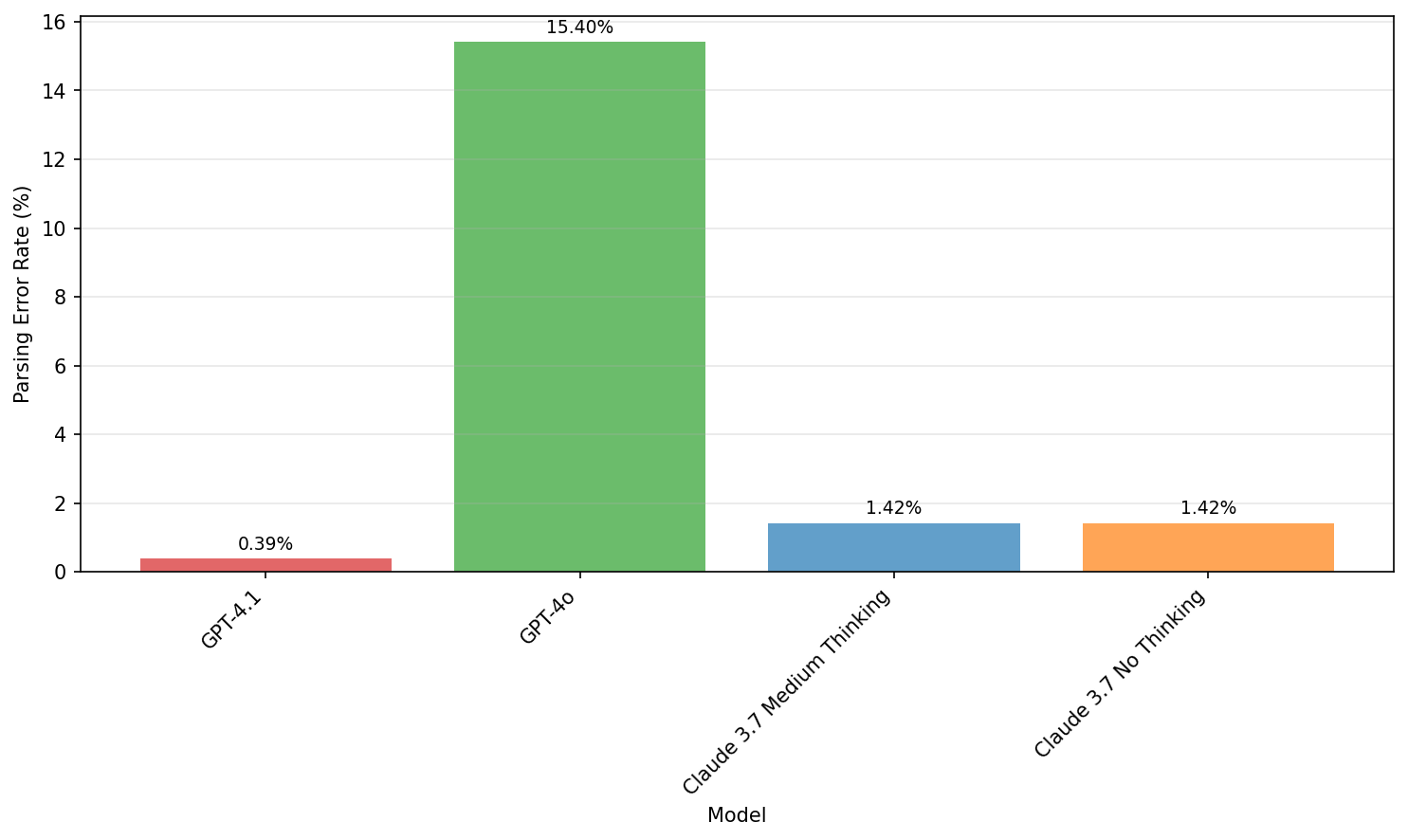}
    \caption{Parsing Error Rates (All tests)}
    \label{fig:parsing_rate}
\end{figure}

\begin{figure}[h!]
    \centering
    \includegraphics[width=0.75\linewidth]{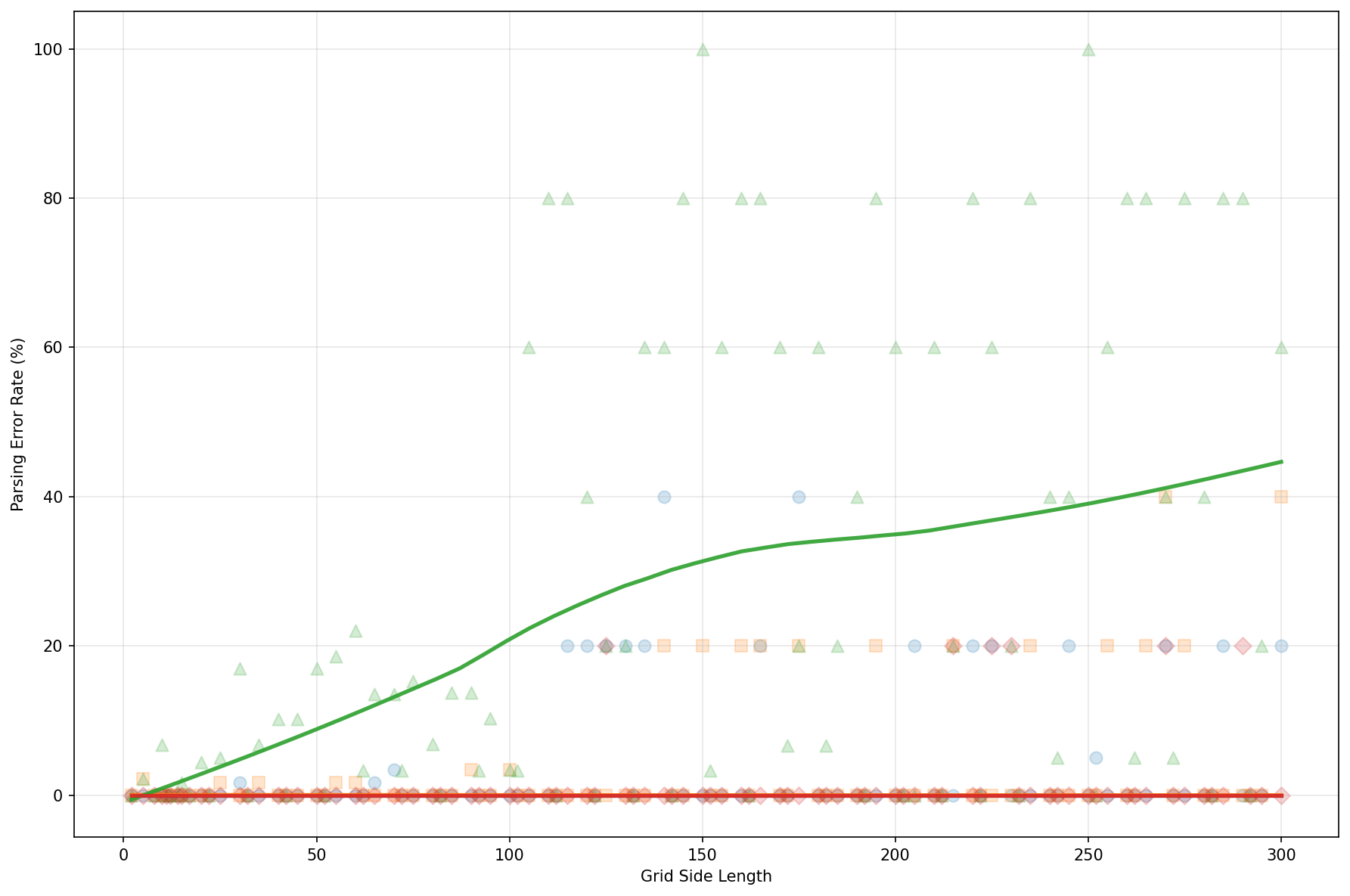}
    \caption{Parsing Error vs Grid Size (All tests)}
    \label{fig:parsing_grid}
\end{figure}

We also encountered some parsing errors with the LLM responses despite efforts with both prompting and regex to make the parsing as reliable as possible. GPT-4o showed significantly higher rates of parsing errors, with 10\% of the total responses in an unparseable or incomplete format. Parsing error rates also increased with grid size, demonstrating once again the further degeneration of model response reliability as complexity increases. These phenomena are seen in Figures~\ref{fig:parsing_rate} and~\ref{fig:parsing_grid}.

\FloatBarrier

\subsection{Tokenization Changes}
In an attempt to mitigate the loss of accuracy, we experimented with several layouts and tokenizations of the grid. Changes were tested on the Quadrant and Search tests. Quadrant was selected because as it was the most internally consistent of the tests, meaning that most problems lay with interpretation of the initial board. Search, on the other hand, was the most difficult, and therefore had the most room for improvement. New grid formats can be checked for tokenization with GPT-4o \href{https://platform.openai.com/tokenizer}{here}, with GPT-4.1 using the same tokenizer. The tokenizer for the Anthropic models is private and therefore unavailable. As such, it should be noted that all references to specific observed tokenizations hold solely for the OpenAI models. 

\subsubsection{Quadrant}
\begin{figure}[h!]
    \centering
    \includegraphics[width=0.9\linewidth]{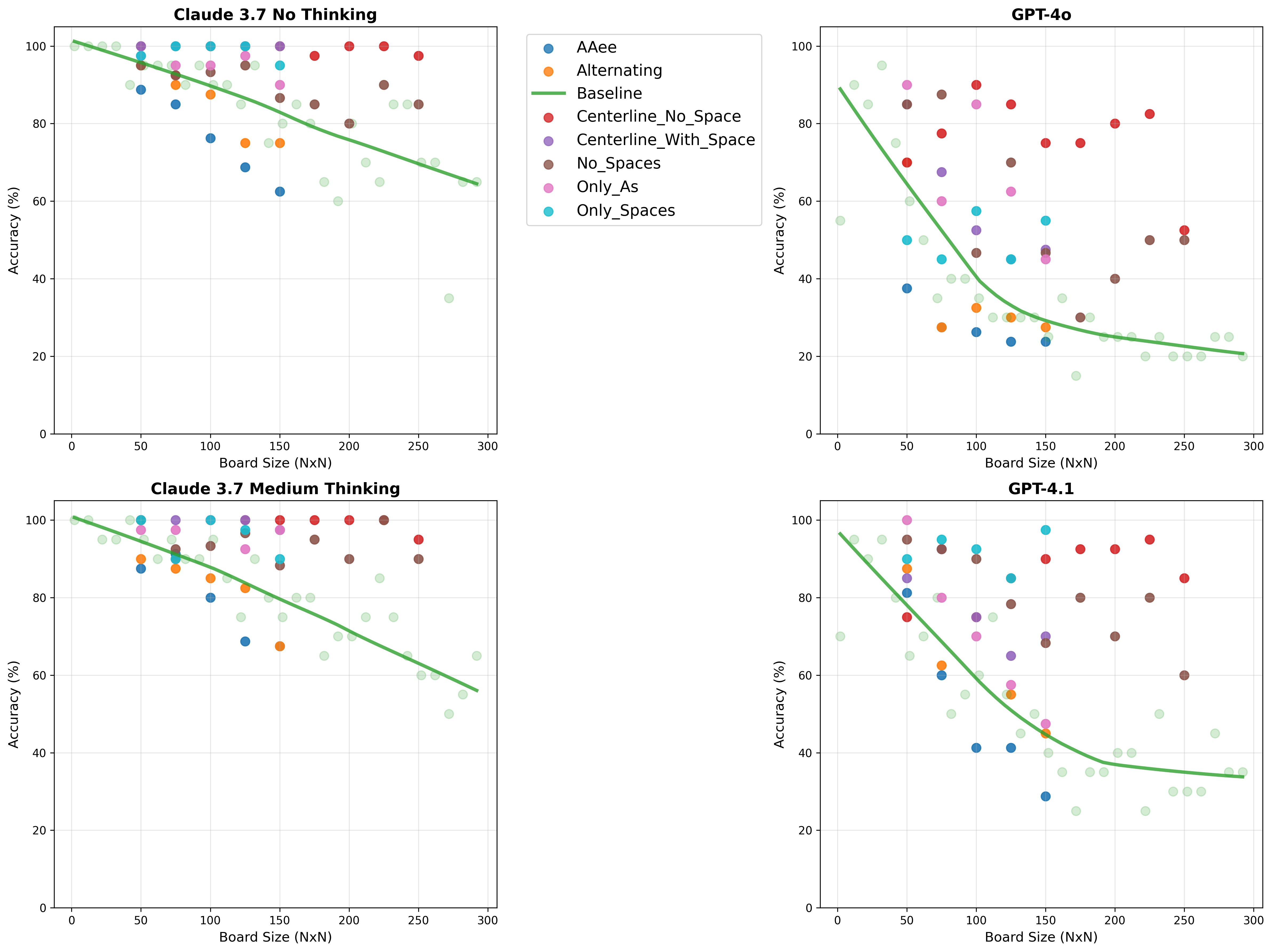}
    \caption{Accuracy vs Grid Size, New Tokenizations (Quad)}
    \label{fig:token_quad}
\end{figure}

The base grid for Quadrant was composed of ``$\cdot$"s separated by spaces, with an X, or other character, taking the place of a ``$\cdot$" where necessary, as seen in the Methodology. The new tests for quadrant can be seen in Figure~\ref{fig:grid_types_quad}, and the results are shown in Figures~\ref{fig:token_quad}.

\begin{figure}[htbp]
    \centering
    
    % Row 1
    \begin{subfigure}{0.3\textwidth}
        \centering
        \raisebox{-0.5cm}{ % Adjust vertical alignment
        $\begin{array}{cccc}
            \cdot & \cdot & \cdot & \cdot \\
            \cdot & \cdot & \cdot & \cdot \\
            \cdot & \cdot & \cdot & \cdot \\
            \cdot & \cdot & \cdot & X
        \end{array}$}
        \caption{No Spaces}
    \end{subfigure}
    \hspace{0.1cm}
    \begin{subfigure}{0.3\textwidth}
        \centering
        \raisebox{-0.5cm}{
        $ \begin{matrix}
            \cdot-\cdot-\cdot-\cdot \\
            \cdot-\cdot-\cdot-\cdot \\
            \cdot-\cdot-\cdot-\cdot \\
            \cdot-\cdot-\cdot-X
        \end{matrix}$}
        \caption{Alternating}
    \end{subfigure}
    \hspace{0.1cm}
    \begin{subfigure}{0.3\textwidth}
        \centering
        \raisebox{-0.5cm}{
        $ \begin{matrix}
            - & - & - & - \\
            | &   & X & | \\
            | &   &   & | \\
            - & - & - & -
        \end{matrix}$}
        \caption{Only Spaces}
    \end{subfigure}
    
    % Row 2
    \vspace{0.3cm}
    \begin{subfigure}{0.3\textwidth}
        \centering
        \raisebox{-0.5cm}{
        $ \begin{matrix}
            AAAA \\
            AAAA \\
            AXAA \\
            AAAA
        \end{matrix}$}
        \caption{Only As}
    \end{subfigure}
    \hspace{0.1cm}
    \begin{subfigure}{0.3\textwidth}
        \centering
        \raisebox{-0.5cm}{
        $ \begin{matrix}
            AAeeAAeeAAeeAA \\
            AAeeAAeeAAeeAA \\
            AAeeAAeeXeeAA \\
            AAeeAAeeAAeeAA
        \end{matrix}$}
        \caption{Alternating AA and ee}
    \end{subfigure}
    \hspace{0.1cm}
    \begin{subfigure}{0.3\textwidth}
        \centering
        \raisebox{-0.5cm}{
        $\begin{matrix}
            \cdot & \cdot & | & \cdot & \cdot \\
            \cdot & \cdot & | & \cdot & \cdot \\
            - & - & + & - & - \\
            \cdot & X & | & \cdot & \cdot \\
            \cdot & \cdot & | & \cdot & \cdot
        \end{matrix}$}
        \caption{Centerline, With Spaces}
    \end{subfigure}
    
    % Row 3
    \vspace{0.3cm}
    \begin{subfigure}{0.3\textwidth}
        \centering
        \raisebox{-0.5cm}{
        
        $\begin{array}{ccccc}
            \cdot \cdot | \cdot \cdot \\
            \cdot \cdot | \cdot \cdot \\
            - - + - - \\
            \cdot \cdot  | \cdot \cdot \\
            \cdot \cdot | X \cdot
        \end{array}
        $}
        \caption{Centerline, No Spaces}
    \end{subfigure}

    \caption{Types of Grids}
    \label{fig:grid_types_quad}
\end{figure}
\FloatBarrier

A few of the grids---Only As, No Spaces, Only Spaces, and both Centerlines---show significant improvement over the baseline. For all tests except the centerline with spaces, a common theme emerges: the ability to tokenize large sections of a line as a singular token. In the baseline grid, each ``$\cdot$" was tokenized individually with the following space, which likely led to confusion and counting problems. With the no spaces grid, large sections could be put into a singular token instead, reducing the counting required and making the ``X" stand out further. The upper bound of tokenization for ``$\cdot$"s appears to be 32 per token, but this is not regular. Smaller chunks than 32 are often split into two tokens despite being under the limit. 

It perhaps seems a moot point to suggest that giving the model the center line improves performance, as it somewhat defeats the purpose of the test. Yet, there are still a few interesting observations to be gleaned from this. First, the addition of centerlines with spaces showed only moderate improvement over the baseline, implying that the models still have a difficult time identifying position in a spaced-out grid, even when given aids for the reasoning portion. Second, though, the increased accuracy implies that the models were using the centerlines as guides. This shows that they are capable of spatial reasoning in a non-strictly mathematical sense, as they could use identifiers like ``left" or ``above" the lines to select the quadrant. Additionally while the Anthropic models average practically 100\% accuracy given the no spaces and centerline, the OpenAI models only average 80\%. While this is a significant improvement, especially on larger boards, it shows that the OpenAI models still have a higher hallucination rate than the Anthropic models, as they still misplace the initial X. 

It is also interesting that the test grid of alternating ``AA"s and ``ee"s performed worse than the baseline. The likely cause of this is that the two sets of characters always tokenize independently of one another, unlike the ``$\cdot$"s that tokenize with the space following. This would lead to the model needing to count and keep track of twice as many entries, resulting in a loss in accuracy due to counting failures. The alternating test also performed worse, even though the grids were tokenized as ``.-" pairs. This is probably due to a lower familiarity with such formats in the training data of the models.

\FloatBarrier

\subsubsection{Search}
\begin{figure}[h!]
    \centering
    \includegraphics[width=1\linewidth]{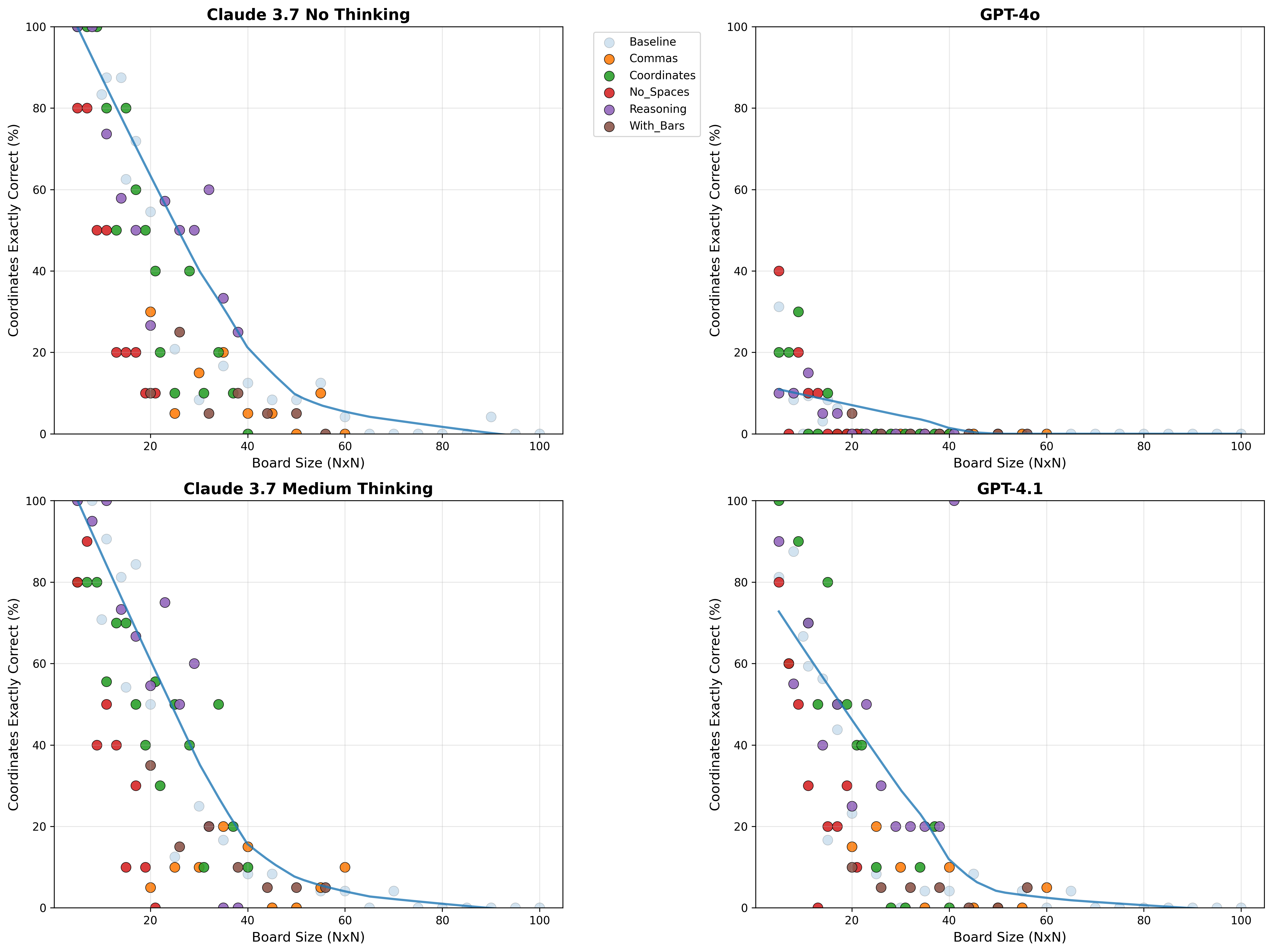}
    \caption{Accuracy vs Grid Size, New Tokenizations (Search)}
    \label{fig:token_search}
\end{figure}

The base grid for Search was letters separated by spaces, as seen in Methodology. The new tests for search can be seen in Figure~\ref{fig:grid_types_search}, and the results can be seen in Figure~\ref{fig:token_search}.

\begin{figure}[htbp]
    \centering
    
    % Force all subfigures to stay in one row with a box
    \mbox{%
    % First subfigure - Commas
    \begin{subfigure}[b]{0.28\textwidth}
        \centering
        % REDUCED height minipage
        \begin{minipage}[c][2.5cm][c]{\textwidth} % Reduced from 4.5cm to 2.5cm
            \centering
            $\begin{array}{cccc}
                A, & B, & C, & D \\
                E, & F, & G, & H \\
                I, & R, & A, & T \\
                M, & N, & O, & P
            \end{array}$
        \end{minipage}
        \caption{Commas}
        \label{fig:commas}
    \end{subfigure}%
    \hspace{0.02\textwidth}%
    
    % Second subfigure - No Spaces
    \begin{subfigure}[b]{0.28\textwidth}
        \centering
        % REDUCED height minipage
        \begin{minipage}[c][2.5cm][c]{\textwidth} % Reduced from 4.5cm to 2.5cm
            \centering
            $\begin{array}{cccc}
                ABCD \\
                EFGH \\
                IRAT \\
                MNOP
            \end{array}$
        \end{minipage}
        \caption{No Spaces}
        \label{fig:no_spaces}
    \end{subfigure}%
    \hspace{0.02\textwidth}%
    
    % Third subfigure - With Bars
    \begin{subfigure}[b]{0.28\textwidth}
        \centering
        % REDUCED height minipage
        \begin{minipage}[c][2.5cm][c]{\textwidth} % Reduced from 4.5cm to 2.5cm
            \centering
            $\begin{matrix}
                - & - & - & - & - & - & - \\
                | & R & | & B & | & C & | \\
                - & - & - & - & - & - & - \\
                | & D & | & O & | & F & | \\
                - & - & - & - & - & - & - \\
                | & G & | & H & | & W & | \\
                - & - & - & - & - & - & -
            \end{matrix}$
        \end{minipage}
        \caption{With Bars}
        \label{fig:with_bars}
    \end{subfigure}%
    } % End of mbox
    
    \caption{Types of Grids}
    \label{fig:grid_types_search}
\end{figure}

\FloatBarrier

We also trialed a non-grid based test, with the grid presented in coordinate form of $(row, column, value)$. Additionally, we tried asking the models to output more thorough step-by-step reasoning before their final answer.

As seen in the results, all tests failed to show any significant improvement over the baseline. The no spaces test, which performed significantly better in the Quadrant trials, showed a much steeper drop in accuracy here. Looking at the tokenization, in the baseline grid, each letter and the following space gets tokenized individually, like in Quadrant. However, without the spaces, the model attempts to form words from the letters, tokenizing them in small chunks. Ironically, even given a word spelled left-to-right, the tokenizer often misses it in favor of smaller groupings. It is interesting as well that step-by-step reasoning has seemingly little effect on performance. The models still skip test cases or miscount letter positions, resulting in a failed identification. These results imply that while LLMs typically excel at language processing, they are still weak in processing in non-traditional formats, such as vertically or diagonally, and are unable to reason their way through the process adequately. 

\FloatBarrier

\section{Discussion}
The results found in the above experiments underscore the fragility and unreliability of LLM spatial reasoning at scale. Across all tests and models, increased grid size led to decreased accuracy and increased error rate. Most models scored high in accuracy on smaller boards, meaning that the tests were not too complex to handle conceptually. Additionally, errors were caused by more than simple coordinate misidentification. For instance, models had difficulty calculating the center lines of the grid in the Quadrant test and made frequent mathematical errors in the Distance test. These types of errors show a more fundamental misunderstanding of spatial reasoning by these LLMs, rather than simply a counting problem. 

This degradation emphasizes the fact that these models are built for linguistic tasks, and, while presenting some emergent behavior \cite{wei2022} \cite{wei2022.2}, are still limited by their architecture. The models possess the reasoning capabilities to handle small examples of a task, but lose track of the grid and task once scaled. 

The differences between the models themselves are just as interesting.
It is immediately evident across the tests that GPT-4o is the weakest at spatial reasoning, followed by GPT-4.1. Both Anthropic models consistently outperformed the OpenAI models, with gaps of up to 30\% in accuracy. GPT-4o also often failed to provide a parseable response despite establishing extensive regex and prompt specifications, whereas none of the other models encountered any such problems.

The error types are interesting as well. The OpenAI models are less internally consistent than the Anthropic ones, displaying weaker consistency in the Quadrant and Transformation tests. In addition, the OpenAI models made more errors in the Distance test, both in position and computation. However, the OpenAI models were less likely to hallucinate a correct answer in the Word Search test. This behavior, of the OpenAI models performing worse in the computational tests, but better in the language-based one, is likely a result of the models' different architectures and training.

\section{Conclusion}

While the Anthropic models outperformed the OpenAI models using this test suite, it is clear that none of the models are particularly good at spatial reasoning. The abilities of the Anthropic models hold promise, but they still deteriorate quickly with scale. We propose a few avenues for future research on this deterioration. For the top models currently released, presentation format and context can significantly affect response quality. For instance, the models tested are all multimodal, and would likely perform better on similar tests presented in image form; future work could focus on directly testing these various modes of input. Other future work could follow up on the changes in grid representation attempted, with an interest in improving efficiency or effectiveness of representations of spatial data in a text-heavy format, in efforts to reduce or minimize the loss in accuracy shown. Further work could also focus on improving the counting and mathematical capabilities of LLMs, as these two skills were the root of many of the inaccurate responses. Finally, research could focus on other methods of output, such as code generation or full boards.

\section*{Acknowledgment}
The authors would like to thank Jump Trading, as well as our mentors Lucas Baker, Nan Yang, Baihong Jin, and Loren Puchalla Fiore.

% \printbibliography[
% heading=bibintoc,
% title={References}
% ]

    \bibliographystyle{icml2024}
% \addbibresource{references.bib}
\bibliography{references}
\end{document}